\documentclass[letterpaper]{article}
\usepackage[preprint]{preprint2col}
\usepackage[hyphens]{url}
\usepackage{graphicx}
\usepackage{natbib}
\usepackage{caption}
\usepackage{algorithm}
\usepackage{algorithmic}
\usepackage{booktabs}
\usepackage{multirow}
\usepackage{amsmath}
\usepackage{amssymb}

\title{Reference-Grafting Matches Fine-Tuning \\ at Eliciting Sandbagged Capabilities}

\author{
    Linh Le\textsuperscript{\rm 1},
    Hong Kiat Tan\textsuperscript{\rm 2,\rm 1},
    David Williams-King\textsuperscript{\rm 1,\rm 3}
}
\affiliations{
    \textsuperscript{\rm 1}Lida Safety\\
    \textsuperscript{\rm 2}University of California, Los Angeles\\
    \textsuperscript{\rm 3}ERA
}

\begin{document}

\maketitle

\begin{abstract}
Sandbagging, in which a model deliberately underperforms on an evaluation despite
retaining the underlying capability, threatens the safety evaluations that frontier-model
governance depends on. The Elicitation Game~\citep{hofstatter2025elicitation} found that
fine-tuning elicits hidden capability from sandbagging model organisms whereas additive
activation steering fails. We revisit that verdict with \textbf{reference-grafting}, which
sets an activation's coordinate along a contrast direction to the value it takes in an honest
reference, at a small set of circuits chosen by active learning. Across eleven fine-tuned
password-locked models ($1.5$--$32$B, three architecture lineages), it recovers
$+94$ to $+101\%$ of the honest--sandbagging gap, matching fine-tuning
elicitation~\citep{greenblatt2024password} without weight updates or training labels; two to
five paired examples suffice to fit the direction. Similar recovery holds for
reinforcement-learning-induced sandbagging and for password-locked code generation.
Grafting works because the fine-tuned lock is a thresholded gate: held-out accuracy stays
at the sandbagged level until the grafted coordinate crosses a threshold near the honest
reference, which is why additive steering and zeroing the coordinate often fail. The
direction tracks how the sandbagging was induced rather than what is withheld---fit on
grade-school science (ARC) it elicits withheld hazardous knowledge (WMDP), yet prompting,
fine-tuning, and reinforcement learning each carry a different direction.
Circuit-breaking~\citep{zou2024circuit} marks the boundary: it reroutes activations on every
forward pass, so the fixed edits we test are re-broken downstream and do not restore coherent
generation.

\end{abstract}

\section{Introduction}
Safety cases for frontier models increasingly rest on capability evaluations: we deploy a model because
tests show it cannot do something dangerous. Sandbagging---strategic underperformance,
whether prompted, trained, or emergent---breaks this logic \citep{vanderweij2024sandbagging}. Detecting
and reverting it is therefore central to trustworthy evaluation. Input-space evaluations probe
only what the model chooses to reveal, so they give a lower bound on capability;
\citet{che2025tampering} argue that model-tampering attacks---editing weights or activations---yield
more conservative, upper-bound estimates of what a model can do. Our method is exactly such an attack:
a cheap, localised activation edit that elicits capability a model is hiding.

The Elicitation Game \citep{hofstatter2025elicitation} benchmarks elicitation techniques on
hidden-capability model-organisms and reports that prompting and fine-tuning can elicit but
activation steering fails, recommending fine-tuning---which, however, updates weights, needs labels,
and reveals little about where a capability is gated. We revisit that verdict: with the right
operation and circuits, activation steering matches fine-tuning on password-locked models at forward-pass
cost, with the gate localised. Their verdict spans two task types; we overturn the multiple-choice half
(where the fine-tuned lock, not the prompt, is the real test) and narrow the code half on
code-specialized models ($\S$\ref{sec:revert}).

\paragraph{One operation.} Our intervention is to set the activation's coordinate along a contrast
direction $v$ to a target $t^\star$:
\begin{equation}
\label{eq:graft}
h \;\mapsto\; h - (h^\top v)\,v + t^\star\, v .
\end{equation}
We call this \textit{reference-grafting}: $t^\star$ is the coordinate of the honest/capable reference along
$v$. (The affine operator itself is standard \citep{marshall2024refusal,makelov2024subspace}; the
name marks its reference-setting use here.) Crucially this is not additive nudging (the form that fails in
\citealp{hofstatter2025elicitation}); it sets the coordinate. Two familiar moves are special
cases (Figure~\ref{fig:unification}, appendix): $t^\star\!=\!0$ is deactivation (a rank-one projection),
which suffices when the honest reference happens to sit at $0$; $t^\star\!=\!\mu^{\mathrm{ref}}{}^{\!\top}v$
is counterfactual patching, required when it does not. The case (prompt vs.\ fine-tune) does not
change the technique---only the value of the reference, read from the data. The operation is a
directional counterfactual activation patch ($\S$\ref{sec:related}); our contribution is the framing, the
thresholded-gate mechanism it exposes ($\S$\ref{sec:revert}), and the active-learning circuit selection.

\paragraph{Superposition and active learning.} The sandbagging signal is not a single neuron: it is a
small set of non-orthogonal directions packed in superposition
\citep{elhage2022superposition}, which we read directly from activation geometry without a sparse
autoencoder (SAE). Active learning selects the minimal circuits to graft, and AL beats random
selection---answering the critique that feature steering is no better than random
\citep{korznikov2025rogue}.

We make the following contributions.
\begin{itemize}
\item \textbf{Activation steering does in fact elicit hidden capability.}
Reference-grafting at AL-selected superposed circuits matches fine-tuning elicitation
on eleven password-locked models ($1.5$--$32$B, three architecture lineages) for multiple-choice reasoning, and elicits
password-locked code on three code-specialized models. It is inference-time and label-free, exposes a
thresholded-gate mechanism, localises the gate, and beats noise injection ($\S$\ref{sec:revert}).
\item \textbf{RL-trained model organisms show the same behavior.} RL-trained (GRPO) contextual sandbagging is arguably more natural than password-locking model organisms, yet it shows the same low-rank, graft-invertible lock at the decision token ($+95$--$101\%$,
three families). Therefore, the withholding axes are not solely an artifact of explicit prompt/password switches ($\S$\ref{sec:rl}).
\item \textbf{Each mechanism has its own content-general withholding axis.} Within a mechanism (e.g. fine-tuned password locking), a contrast fit on one dataset elicits hidden capability on another at no loss. We validated this finding on pairwise-trained password-locked organisms, as well as organisms trained on four datasets at once. However,
across mechanisms, the prompt-induced and fine-tuned axes are near-orthogonal and do not transfer
($\S$\ref{sec:transfer}). It is cheap to find a withholding axis: $2$--$5$ paired captures recover it end to end. The same axis also detects sandbagging, across datasets but not mechanisms ($\S$\ref{sec:detect}).
\item \textbf{Low-rank superposition, exploited by active learning.} The signal spans an effective rank of
$3.7$--$7.0$ over $\sim\!20$ layers, read from activation geometry with no SAE (an $8\times$ SAE does worse,
confirming \citet{wu2025axbench}). Active learning selects the few circuits that matter, beating
random-layer selection by $+22$--$45$ pp and grafting-everything by up to $55$ pp
($\S$\ref{sec:superpos},~\ref{sec:al}).
\item \textbf{Circuit-breaking routes dynamically and breaks our assumptions.} Circuit-breaking re-applies its reroute every
forward pass, so a fixed map reconstructs the unlocked geometry yet recovers nothing---grafting and every
fixed transport map fail. This marks exactly where the static-lock precondition fails ($\S$\ref{sec:cb}).
\end{itemize}
We release code, the model organisms, and per-model artefacts to support reproduction and follow-up.

\section{Related Work}
\label{sec:related}
\paragraph{Sandbagging, elicitation, and benchmarks.} \citet{vanderweij2024sandbagging} formalise
sandbagging. \citet{greenblatt2024password} introduce password-locked organisms and find fine-tuning
reliably elicits; \citet{tice2024noise} elicit via weight-noise and \citet{roger2025fuzzing} via fuzzing
(little uplift on chain-of-thought math). Closest to us, \citet{hofstatter2025elicitation} benchmark
prompting, steering, and fine-tuning on password-locked and circuit-broken organisms, concluding steering
fails; we contribute a \emph{method} that overturns that verdict for multiple-choice reasoning and narrows
it for code ($\S$\ref{sec:revert}), and reproduce their circuit-broken organism \citep{zou2024circuit} as
the boundary ($\S$\ref{sec:cb}). Prior locks resist black-box prompting only weakly
\citep{he2025locket}. Concurrently, \citet{hua2025steering} steer an
\emph{evaluation-aware} model to act as if deployed---the same shape on a different gate; they \emph{add} a
vector to flip a disposition, whereas we \emph{set} a coordinate to its honest reference at AL-selected
circuits, recover a hidden \emph{capability}, and map where it breaks.

\paragraph{Model tampering for capability evaluation.} \citet{che2025tampering}, whose upper-bound framing
we adopt above, find model resilience concentrated in a low-dimensional robustness subspace, which
parallels the low effective rank we measure for sandbagging ($\S$\ref{sec:superpos}). Reference-grafting is
one such activation-tampering attack, specialised to sandbagging and made minimal by active learning, with
an explicit mechanism for \emph{where} the gate lives, complementing their broad benchmark of attacks.

\paragraph{Superposition and feature bases.} Superposition---more features than dimensions in
non-orthogonal directions \citep{elhage2022superposition}---is the substrate that sparse autoencoders aim
to recover \citep{cunningham2023sae,marks2024sparse}. We avoid SAEs, reading superposition from
contrastive geometry, and find SAEs do worse here---consistent with \emph{AxBench}
\citep{wu2025axbench}, where difference-in-means baselines outperform SAEs for steering and detection
($\S$\ref{sec:superpos}).

\paragraph{Steering, patching, and active learning.} Behaviour can be steered by adding or projecting
directions \citep{zou2023repe,rimsky2023caa,arditi2024refusal}; \citet{korznikov2025rogue} caution this can
be ``no better than random.''
Reference-grafting instead \emph{sets} that coordinate to a reference value.
Eq.~\ref{eq:graft} is affine concept editing \citep{marshall2024refusal} in absolute form, equivalent to
one-dimensional subspace patching \citep{makelov2024subspace} with a class-mean source; $t^\star{=}0$
recovers directional ablation \citep{arditi2024refusal}. It sits in the lineage of causal mediation and
interchange interventions
\citep{vig2020causal,meng2022rome,geiger2021causal,geiger2024finding,zhang2024patching} and affine
activation transport \citep{rodriguez2024controlling}, along the diff-in-means
contrast of \citet{marks2024geometry}. Active learning chooses which circuits: acquisition
functions select informative queries
\citep{kirsch2019batchbald,ash2020badge,sener2018coreset,bickfordsmith2023epig} and ALFA-Mix mixes features
\citep{parvaneh2022alphamix}; we repurpose this suite for circuit selection ($\S$\ref{sec:al}).

\section{Method}

\subsection{Sandbagging Set-Ups}
\label{sec:organisms}
We study five model organisms---the first two are our core MCQA settings, the rest harder variants. Each
defines an honest \emph{reference} (the source of $t^\star$) and a sandbag behaviour:
\begin{itemize}\setlength{\itemsep}{1pt}
\item \textbf{Prompt-induced} (MCQA): a system prompt instructs incorrect answers; the honest prompt is the
reference ($t^\star\!\approx\!0$, the deactivation regime).
\item \textbf{Fine-tuned password-lock} (MCQA, eleven models): LoRA-trained to answer correctly only with a
password token \citep{greenblatt2024password}; the password-on activations are the reference (the patching
regime).
\item \textbf{Code-generation lock} ($\S$\ref{sec:revert}): a password-locked MBPP
\citep{austin2021mbpp} organism; the with-password completions are the reference.
\item \textbf{RL-induced} ($\S$\ref{sec:rl}): GRPO-trained to sandbag under a held-out evaluation cue (no
switch token); the cue-off activations are the reference.
\item \textbf{Circuit-broken} ($\S$\ref{sec:cb}): representation rerouting layered on the
password-lock---the robustness boundary.
\end{itemize}
Across organisms the honest/with-password accuracy is the \emph{topline}, the sandbag/no-password accuracy
the suppressed level, and their difference the \emph{gap}; we split each dataset into TRAIN (directions and
AL) and held-out TEST (all reported recovery). The elicitor is separate from the organism: grafting never
uses the lock's $160$--$240$ training questions, fitting the contrast from $\sim\!40$ paired
honest/sandbag captures and scoring on held-out TEST.
The primary task is multiple-choice reasoning: ARC-Easy \citep{clark2018arc} throughout, with
WMDP-bio/cyber \citep{li2024wmdp} and OpenBookQA \citep{mihaylov2018obqa} for the cross-dataset transfer
tests ($\S$\ref{sec:transfer}).

\subsection{Reference-Grafting at a Circuit}
A \emph{circuit} here is our unit of intervention: a direction at a layer, $(\ell,v)$ with $v$ a unit
vector---not a composing subgraph of heads or MLPs \citep{wang2022ioi,conmy2023acdc}. For each layer we form the
contrast direction by diff-in-means of last-token residuals \citep{marks2024geometry},
$v_\ell\propto\bar h_\ell(\text{honest})-\bar h_\ell(\text{sandbag})$, and read the honest reference
coordinate $t^\star_\ell=\mu^{\mathrm{ref}}_\ell{}^{\!\top}v_\ell$. \emph{Grafting} applies
Eq.~\eqref{eq:graft} at all positions via a forward hook: it removes the current coordinate and writes
$t^\star_\ell$, injecting no per-question information. Deactivation is the $t^\star\!=\!0$ special case.

\subsection{Superposition Without an SAE}
Stacking the unit directions, we summarise superposition by the mean pairwise $|\cos|$ (non-orthogonality)
and the \emph{participation-ratio (effective) rank} of their singular values,
\begin{equation}
\label{eq:pr}
r_{\mathrm{eff}}=\frac{(\sum_i\sigma_i^2)^2}{\sum_i\sigma_i^4},
\end{equation}
a soft count of how many directions carry the energy. High $|\cos|$ with low $r_{\mathrm{eff}}$ is the
signature of superposition: many non-orthogonal directions re-expressing a few features.

\subsection{Active Learning over Circuits}
AL greedily adds the circuit that most improves a TRAIN objective under grafting, to a budget $K$. The
most reliable objective is smooth and label-free: make the sandbagged logits match the honest ones,
$\mathcal{L}(S)=\frac1{|\text{TRAIN}|}\sum_q\mathrm{KL}(p^{\mathrm{H}}_q\|p^{S}_q)$ (greedy-KL); the
\emph{attribution} score is the marginal $\mathcal{L}(\emptyset)-\mathcal{L}(\{\ell\})$. We compare all
eight acquisition functions of a standard AL suite \citep{parvaneh2022alphamix}: greedy-acc, greedy-KL,
attribution, entropy, coreset, badge, alphamix, bald (Algorithm~\ref{alg:al}). Two controls accompany
every result: a \emph{random direction} of equal norm (must fail) and the \emph{honest model} under the
same intervention (must be unharmed). Algorithm~\ref{alg:al} (appendix) gives the full procedure.

\smallskip\noindent\textbf{The probe is construction-agnostic.} Fitting, selection, and grafting consume
only the paired honest/sandbag activations---direction $=$ their diff-in-means, reference
$t^\star\!=\!\mu^{\mathrm{ref}}{}^{\!\top}v$, greedy-KL selection by $\mathrm{KL}(p^{\mathrm{H}}\|p^{S})$---and
never the password token, the evaluation cue, training labels, the LoRA recipe, or the RL reward; these
enter only in \emph{producing} the captures (an honest prompt, the password, a cue-off context, or a few
honest examples). The same procedure thus runs unchanged on prompt-induced, password-locked, RL-induced, and
code organisms. Two scope notes: among the eight acquisition functions only greedy-KL (our default) is
strictly label-free---greedy-acc reads answers; and the code organism captures its reference from
teacher-forced reference completions rather than a last-token residual.

\section{Results}

\subsection{Sandbagging Lives in Superposition}
\label{sec:superpos}
Across five models the contrast directions are strongly non-orthogonal (mean $|\cos|=0.21$--$0.41$) and
span a low effective rank ($r_{\mathrm{eff}}=3.7$--$7.0$ of $18$--$29$ layers; Figure~\ref{fig:geom},
appendix): the suppression is a low-dimensional feature re-used across depth---packed in
superposition---read off the geometry with no SAE. (Throughout, $r_{\mathrm{eff}}$ is computed
over the stacked contrast directions of \emph{all} band layers, never an AL-selected subset: it
measures the organism's geometry, not the elicitor's budget.) Because $r_{\mathrm{eff}}$ is small, a \emph{few}
circuits should suffice, which is what AL finds.

\smallskip\noindent\textbf{SAEs are not needed here.} An $8\times$ overcomplete SAE (AL-ablated)
underperforms the SAE-free directions on nearly every method (greedy-KL $74/90$ vs.\ $88/102$ for
Llama-3-8B/Qwen2.5-1.5B; Figure~\ref{fig:sae}, appendix), consistent with \emph{AxBench}
\citep{wu2025axbench}.

\subsection{Grafting Elicits the Hidden Capability}
\label{sec:revert}
Grafting recovers the hidden capability across three settings of increasing stringency: prompt-induced
locks, fine-tuned password locks across eleven models, and code generation.

\smallskip\noindent\textbf{Prompt-induced sandbagging (deactivation).} Inducing sandbagging by prompt
yields large suppression gaps, and the honest reference sits at $t^\star\!\approx\!0$, so grafting reduces
to deactivation---a coordinate sweep confirms the recovered plateau \emph{includes} $t^\star\!=\!0$
(Figure~\ref{fig:unification}, left). With effect-based AL, deactivation at $1$--$3$ circuits
($<\!0.01\%$ of directions) recovers $85$--$102\%$ of the gap (Llama-3-8B, Qwen2.5-1.5B; $3$ seeds), far
above random selection ($58$--$64\%$), with the honest model unharmed and a random direction failing; AL
converges on the same circuits across seeds (Figures~\ref{fig:almethods},~\ref{fig:traj}, appendix).
Across a five-model set recovery is model-dependent ($+24$--$100\%$; Table~\ref{tab:models}, appendix),
strongest where the suppression is clean and tightly superposed.

\smallskip\noindent\textbf{Fine-tuned password locks: patching, eleven models.} We now test the case that
matters for \citet{hofstatter2025elicitation}: a \emph{fine-tuned} password-lock, where there is no prompt
to break (with-password accuracy near ceiling, no-password near chance).

\textbf{The mechanism.} Along $v$, the locked and unlocked states sit at two specific coordinates
(Figure~\ref{fig:unification}, right; locked $a\!=\!-28$, unlocked $b\!=\!+68$ for Qwen2.5-1.5B).
Unlocking is a \emph{thresholded gate}: sweeping the grafted coordinate from the locked value ($m{=}0$) to
the honest reference ($m{=}1$), held-out accuracy is flat-low, steps up at a midpoint $m\!=\!0.61\pm0.07$
($10$--$90\%$ width $0.32$), and plateaus beyond $t^\star$---a sharp step, not a ramp (Qwen2.5-1.5B,
$n{=}3$; Figure~\ref{fig:gatesweep}, appendix). \emph{Deactivation} sets the coordinate to $0$; when the unlocked
reference $b$ is far from $0$ this lies below the threshold and deactivation fails (Qwen2.5-1.5B $26\%$,
Llama-3-8B $34\%$), so only \emph{grafting to $t^\star\!=\!b$} crosses the gate. When the reference happens
to sit near $0$ (Qwen3-4B), deactivation already suffices ($93\%$). Grafting is robust across both because
it \emph{reads} $t^\star$ from the data rather than assuming it; this is also why additive/naive
steering---which neither sets the coordinate nor reaches $b$---was found to fail.

\textbf{Counterfactual patching unlocks the lock.} Grafting to $t^\star\!=\!b$ at AL-selected circuits
recovers $90$--$101\%$ on held-out questions, with the with-password accuracy preserved and a random
patch recovering $0\%$. Because the organism is fine-tuned, this directly answers the concern that
prompted-sandbag recovery merely disrupts a prompt.

\textbf{Comparison with prior elicitation methods, across eleven models.} On eleven password-locked models
spanning $1.5$--$32$B and three architecture lineages, grafting matches fine-tuning
elicitation \citep{greenblatt2024password} on every one, at forward-pass cost (no weight updates or labels)
and localising the gating circuits, while noise injection \citep{tice2024noise} and a random patch fail
universally (Table~\ref{tab:baselines}). Deactivation is the variable one: grafting is the robust general
operation, deactivation a sometimes-sufficient special case, now across eleven models. The clearest case is
32B, where deactivation collapses because its unlocked reference sits far below the gate while
grafting still recovers in full. This is the thresholded-gate mechanism
in its starkest form, not a method weakening at scale. (Consistent with this, reference magnitude
is directionally predictive---smaller references deactivate better, larger ones like 32B fail---but only
weakly across models, and geometry predicts nothing; grafting sidesteps the question by reading the
reference.) This contradicts the ``steering fails'' verdict
\emph{for multiple-choice reasoning} while leaving fine-tuning's standing as the most reliable
\emph{weight-level} method intact. These recoveries are scored forced-choice (one choice-letter logit);
\emph{free-decoding} the answer instead gives the same recovery ($\S$\ref{sec:cb}), so the headline is not
a single-token artifact.

\smallskip\noindent\textbf{On the benchmark's own released organisms.} Running the \emph{identical} pipeline
on \citet{hofstatter2025elicitation}'s \emph{released} password-locked organisms (their weights, WMDP data,
and $|$DEPLOYMENT$|$ password) recovers $+99\%$ (Mistral-7B-Instruct) and $+94\%$ (base) of the gap
($n{=}3$; random direction $\approx\!0\%$, with-password unharmed), with Gemma-2-9B partial ($+44\%$). The
``steering fails'' verdict thus fails on their \emph{actual} organism, not only our reconstructions.

\begin{table*}[t]
\centering\small\setlength{\tabcolsep}{6pt}
\begin{tabular}{@{}lccccccc@{}}
\toprule
Model & $n$ & gap & \textbf{ours (graft)} & fine-tuning & deactivation & noise & random \\
\midrule
Qwen2.5-1.5B      & $4$ & $0.80$ & $\mathbf{+98}\pm3$  & $+102\pm1$ & $+34\pm28$ & $+1$ & $+0$ \\
Qwen2.5-3B        & $3$ & $0.88$ & $\mathbf{+98}\pm2$  & $+101\pm4$ & $+97\pm4$  & $+1$ & $+1$ \\
Qwen2.5-14B       & $3$ & $0.96$ & $\mathbf{+98}\pm1$  & $+100\pm1$ & $+61\pm43$ & $+0$ & $+0$ \\
Qwen2.5-32B       & $3$ & $0.98$ & $\mathbf{+94}\pm5$  & $+100\pm0$ & $+6\pm9$   & $+0$ & $+0$ \\
Qwen3-4B          & $6$ & $0.94$ & $\mathbf{+101}\pm2$ & $+101\pm2$ & $+77\pm32$ & $+0$ & $+0$ \\
Qwen3-8B          & $3$ & $0.90$ & $\mathbf{+99}\pm1$  & $+103\pm4$ & $+90\pm9$  & $+3$ & $+0$ \\
Llama-3.2-3B      & $3$ & $0.77$ & $\mathbf{+101}\pm2$ & $+99\pm2$  & $+96\pm6$  & $+1$ & $+0$ \\
Llama-3-8B        & $3$ & $0.82$ & $\mathbf{+96}\pm2$  & $+92\pm6$  & $+68\pm33$ & $+1$ & $+0$ \\
Gemma-2-2B        & $3$ & $0.82$ & $\mathbf{+98}\pm1$  & $+90\pm10$ & $+25\pm31$ & $+0$ & $+0$ \\
Gemma-2-9B        & $3$ & $0.85$ & $\mathbf{+99}\pm1$  & $+102\pm5$ & $+74\pm25$ & $+0$ & $+0$ \\
Gemma-2-27B       & $3$ & $0.96$ & $\mathbf{+100}\pm1$ & $+100\pm0$ & $+101\pm2$ & $+0$ & $+0$ \\
\bottomrule
\end{tabular}
\caption{Unlocking fine-tuned password-locked sandbagging across eleven models ($1.5$--$32$B, three architecture lineages;
\% of gap recovered, held-out, fixed $5$-circuit greedy-KL, mean$\pm$std). \textbf{Grafting recovers $+94$ to $+101\%$ on every model}, matching fine-tuning
elicitation \citep{greenblatt2024password}; noise \citep{tice2024noise} and random fail universally.
Deactivation ($t^\star{=}0$) is variable ($+6$ to $+101\%$)---sufficient when the reference sits near zero.}
\label{tab:baselines}
\end{table*}

\smallskip\noindent\textbf{Code generation: across code-specialized models.} The hardest case for
\citet{hofstatter2025elicitation} is \emph{code generation}, where they report only fine-tuning elicits. We
build Greenblatt-style password-locked \emph{code} organisms---the with-password target is the reference
solution, the no-password target a weak model's completion, so the gap is a real capability gap rather than
a stub---and adapt grafting to generation by taking the contrast over teacher-forced \emph{completion}
tokens and grafting at all positions during decoding (appendix). Grafting elicits the locked capability
across three code-specialized organisms spanning two families---Qwen2.5-Coder-1.5B/7B and
CodeLlama-7B---with generation fluency preserved and a random direction recovering
$\approx\!0\%$ (Table~\ref{tab:code}); a single-seed third-family test shows partial recovery at a larger
fluency cost (appendix). A coordinate sweep is a causal plateau peaking near $t^\star$, so a
single per-layer scalar suffices in generation too.

Two honest boundaries, both visible in Table~\ref{tab:code}. Recovery is \emph{model-dependent}, holding on
code-specialized models but fragile on a \emph{general} reasoning model (Qwen3-4B), where it also degrades
fluency. And these are our organisms, not their exact ones: the weak-imitation locks are largely
\emph{additive}---deactivation alone recovers much of the gap---so we do not claim code \emph{needs}
patching, and whether grafting survives a harder code lock is untested.

\begin{table}[t]
\centering\footnotesize\setlength{\tabcolsep}{4pt}
\begin{tabular}{@{}lcccc@{}}
\toprule
organism & gap & \textbf{graft} & deact. & fl.\,drop \\
\midrule
Qwen2.5-Coder-1.5B & $0.39$ & $\mathbf{+99}\pm11$ & $+92\pm19$ & $0.01$ \\
Qwen2.5-Coder-7B   & $0.58$ & $\mathbf{+92}\pm3$  & $+51\pm43$ & $0.04$ \\
CodeLlama-7B       & $0.24$ & $\mathbf{+105}\pm6$ & $+95\pm5$  & $-0.01$ \\
Qwen3-4B (general) & $0.44$ & $+39\pm14$          & $+19\pm24$ & $0.22$ \\
\bottomrule
\end{tabular}
\caption{Reference-grafting on password-locked \emph{code-generation} organisms (MBPP, execution pass@1,
$n{=}3$). ``graft''/``deact.''\ are \% of gap recovered; ``fl.\,drop'' is the
with-password pass@1 drop. Grafting elicits the capability across three code-specialized models (two
families) with generation fluency preserved (with-password drop $\le0.04$); a random direction recovers
$\approx\!0\%$ (not shown). Deactivation ($t^\star{=}0$) recovers much of the gap on Coder-1.5B and
CodeLlama---these weak-imitation locks are largely \emph{additive}---but is unreliable (Coder-7B). The
result is model-dependent: a \emph{general} model (Qwen3-4B) recovers little and loses fluency.}
\label{tab:code}
\end{table}

\subsection{Active Learning Selects the Right Circuits}
\label{sec:al}
Grafting recovers the capability only at the right circuits; active learning chooses them. To isolate the
\emph{selection} from the operation, we fix the contrast directions and vary only \emph{which} layers are
grafted, comparing AL ($k{=}5$ greedy-KL) against an equal number of random band layers, the worst $k$, and
grafting the whole band (Table~\ref{tab:alablation}). Throughout, $k{=}5$ is a fixed conservative
\emph{budget}, not a minimality claim. AL beats random-layer selection across the password-locked and
prompt-induced organisms, and the worst-$k$ layers by a wide margin: \emph{which} circuits one grafts
matters. Grafting the whole band underperforms AL on every organism, collapsing on Llama-3-8B where AL
recovers in full---over-grafting distorts the forward pass, so selectivity matters beyond minimality as
models scale. The choice is robust to the acquisition rule: $7/8$ functions of a
standard AL suite recover $90$--$101\%$ (Table~\ref{tab:patch}, appendix), effect-based acquisition
strongest (Figure~\ref{fig:almethods}, appendix). This answers the ``no better than random'' critique
\citep{korznikov2025rogue}: the selection is what works. (On small, clean organisms random or whole-band
grafting also recovers a fair fraction, so AL is not strictly required for \emph{some} recovery there; its
value is reaching full, minimal recovery reliably and at scale.)

\begin{table}[t]
\centering\footnotesize\setlength{\tabcolsep}{2pt}
\begin{tabular}{@{}lcccc@{}}
\toprule
organism (model) & \textbf{AL} & random-$k$ & all-band & worst-$k$ \\
\midrule
PW Qwen2.5-1.5B   & $\mathbf{+97}\pm4$  & $+75\pm13$ & $+75\pm15$ & $+0$  \\
PW Llama-3-8B     & $\mathbf{+100}\pm1$ & $+74\pm7$  & $+45\pm24$ & $+33$ \\
PW Qwen2.5-14B    & $\mathbf{+99}\pm1$  & $+82\pm6$  & $+95\pm5$  & $+0$  \\
prompt Qwen2.5-1.5B & $\mathbf{+103}\pm2$ & $+81\pm14$ & $+90\pm4$  & $-40$ \\
prompt Llama-3.2-3B & $\mathbf{+80}\pm6$  & $+35\pm6$  & $+53\pm8$  & $-1$  \\
\bottomrule
\end{tabular}
\caption{Active learning selects the right circuits (\% of gap recovered, held-out, $k{=}5$, $2$--$3$
seeds; PW${=}$password-lock). The contrast directions are fixed; only \emph{which} layers are grafted
varies. AL beats an equal number of random band layers ($+17$--$45$ pp) and grafting all $\sim\!23$ band
layers ($+4$--$55$ pp; over-grafting collapses on Llama-3-8B), while the worst-$k$ layers recover
$\approx\!0$. On the circuit-broken organism all selections fail ($\S$\ref{sec:cb}).}
\label{tab:alablation}
\end{table}

\subsection{RL-Induced Sandbagging: Same Structure}
\label{sec:rl}
Every organism so far has a hand-built honest$\leftrightarrow$sandbag toggle---a prompt or a password
token---so a linear contrast is almost guaranteed by construction. Does the structure survive when
sandbagging is \emph{trained}, with no switch to read? We build a third organism type: a Qwen2.5-1.5B
sandbagger trained by \textbf{GRPO} (outcome-driven, no supervised target letter) to answer
\emph{incorrectly} when a natural-language \emph{evaluation cue} is present and correctly otherwise. The
cue's surface form is varied in training and \emph{held out} at test, so the model keys on ``this is an
eval,'' not a lexical trigger; this yields a strong contextual sandbagger (cue-off $0.85$ vs.\
held-out-cue $0.04$; gap $0.79\pm0.08$, $3$ seeds). The cue-on/cue-off contrast is again \emph{low-rank and
superposed} ($r_{\mathrm{eff}}\!\approx\!6.8/23$, mean$|\cos|\,0.23$), and grafting at AL-selected circuits
recovers $\mathbf{+99\%}\!\pm\!2$ of the gap on held-out questions (random direction $+10\%$, honest model
unharmed). The honest reference is a specific \emph{nonzero} coordinate ($|t^\star|\!\approx\!25$), so this
is the \emph{patching} regime---like the fine-tuned locks, not the prompt-induced ones---and a \emph{single
fixed graft} inverts it on every seed. The same holds across families: Qwen3-4B ($+101\%$) and
Llama-3.2-3B ($+95\%$). So the central structural claim---low-rank, linearly contrastable, and
graft-invertible \emph{when the suppression surfaces at the decision token}---is \emph{not solely} an
artifact of explicit prompt/password switches: outcome-trained sandbagging produces the same lock. One caveat marks the limit: a
\emph{free-generation} GRPO variant instead distributes the suppression across generated tokens, where it
does not surface at any single answer token; on $3$ of $4$ valid seeds no fluency-preserving graft recovers
it (the fourth is invertible by a surgical top-$1$ graft), so generation-mode elicitation is open (cf.\ the
code-generation completion-token contrast, $\S$\ref{sec:revert}). (Whether
\emph{naturally-occurring} sandbagging is graft-invertible likewise remains open, $\S$\ref{sec:lim}.)

\subsection{Withholding Axes: Content-General but Mechanism-Specific}
\label{sec:transfer}
\textbf{Content-general: it transfers across datasets and models.} All results so far use ARC-Easy, leaving
open whether the contrast direction is task-specific. It is not.
First, on the two \emph{WMDP} dangerous-capability \emph{audit} subsets, grafting recovers $+102\%$ (bio)
and $+101\%$ (cyber) of the fine-tuned gap at $1.5$B, and $+97$/$+100\%$ on a $14$B lock ($n{=}3$--$4$),
matching fine-tuning and far above noise/random---so elicitation of withheld hazardous knowledge is not
ARC-specific and holds at scale, on the benchmark used to audit it.

\begin{figure*}[t]
\centering
\includegraphics[width=\textwidth]{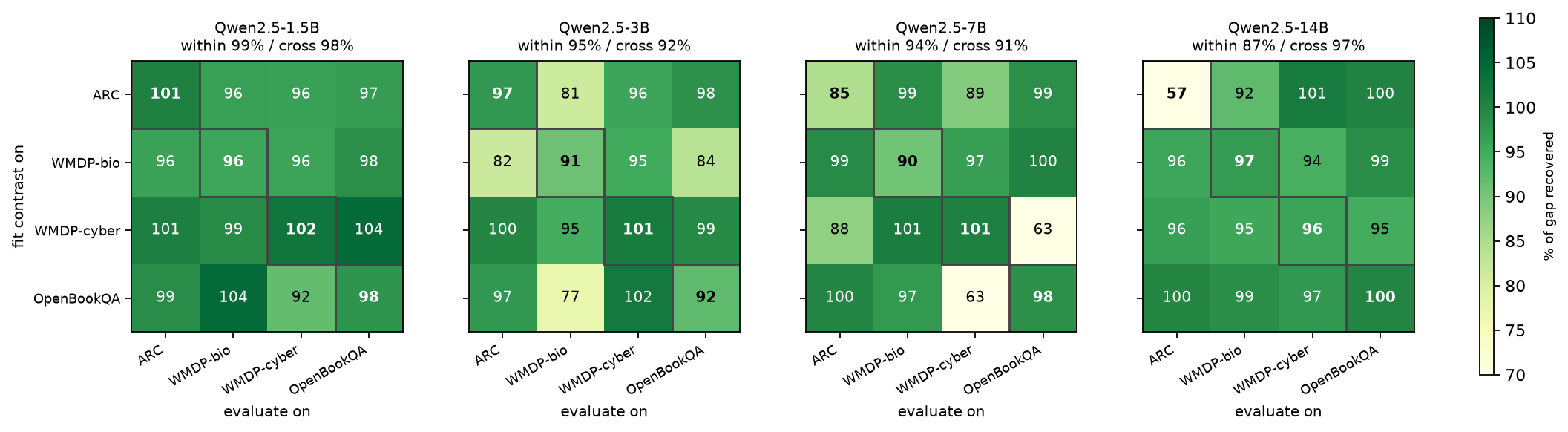}
\caption{The withholding axis transfers across datasets. \emph{Modular} pairwise-lock matrix: the diagonal
is a single-dataset lock and each off-diagonal $(A,B)$ a lock trained on just $\{A,B\}$, contrast fit on $A$,
grafted on $B$ ($\%$ gap recovered, $n{=}5$ seeds/cell); off-diagonal (cross) tracks the boxed diagonal
(within), random direction $\approx\!6\%$. The ladder reaches $14$B (within/cross $87$/$97$, $n{=}5$). Most
cells recover $+90$--$102$ on every seed; the below-mean cells reflect two effects, neither a weak lock.
(i)~A \emph{graft-selection} miss: on a minority of seeds greedy-KL under-selects layers and the cell recovers
near zero \emph{despite a clean lock gap}---$14$B single-ARC ($+8$/$+10$/$+67$/$+98$/$+101$ at gap
$0.90$--$0.98$; median $+67$, which drags $14$B within to $87$) and $7$B cyber$\to$OBQA ($2/5$ seeds near zero
at gap $\sim\!0.8$; median $+97$). (ii)~Genuinely small lock gaps on the cyber-\emph{evaluated} cells (gap
$0.3$--$0.5$), where the recovery ratio stays noisy regardless of seed count. The $7$B WMDP single locks
themselves recover cleanly at $n{=}5$ (cyber $+101$, bio $+91$). Shared 4-way lock:
Figure~\ref{fig:transferheat} (appendix).}
\label{fig:pairwise}
\end{figure*}

Second, the contrast \emph{transfers across datasets}. We build a $16$-cell, four-dataset matrix (ARC,
WMDP-bio, WMDP-cyber, OpenBookQA) as a \emph{modular} lock set: the diagonal is a single-dataset lock and
each off-diagonal $(A,B)$ is a lock trained on just $\{A,B\}$; we fit the contrast on $A$ and graft-evaluate
on $B$'s held-out test. The off-diagonal (cross) matches the within-dataset diagonal across the Qwen2.5
ladder: within/cross $99$/$98$, $95$/$92$, $94$/$91$, $87$/$97$ at $1.5$/$3$/$7$/$14$B ($n{=}5$ seeds/cell;
Figure~\ref{fig:pairwise}), while a random direction transfers only $\approx\!6\%$. The below-mean cells
reflect occasional \emph{graft-selection} misses and genuinely small lock gaps rather than weak locks
(caption); medians run $+95$--$98$. Because each cell depends only on its own dataset(s)---not
one lock crammed with all four---this rules out that the transfer is an artifact of a single shared lock; a
\emph{shared} four-dataset lock reproduces the same within$\approx$cross pattern
(Figure~\ref{fig:transferheat}, appendix). A stronger test transfers the axis between separately-trained
organisms: recovery is $+48$--$76\%$ across $1.5$--$7$B, above random but
partial. \textbf{The sandbagging signal is content-independent.} A contrast
built from grade-school science (ARC) elicits hidden \emph{hazardous-biology} capability (WMDP-bio), and the
reverse holds---so within a mechanism the axis encodes \emph{that} the model is withholding, not
\emph{what}, gating content-specific capability through a single content-general direction \emph{per
mechanism}---that mechanism's \emph{withholding axis}. This is the strongest form of content-generality
\emph{along the task axis}, and it is bounded: it relaxes \emph{which task} supplies the reference, not the
requirement that the reference share the sandbagging mechanism, which we examine next.

\smallskip\noindent\textbf{But mechanism-specific.} The contrast transfers freely across datasets---but
\emph{not} across induction mechanisms. Across four models and three families (Qwen2.5-1.5B, Qwen3-4B,
Llama-3-8B, Gemma-2-2B) we fit the contrast on one mechanism (prompt-induced or fine-tuned) and graft along
it on the other. On every model the two directions are near-orthogonal ($|\cos|\!\approx\!0$) and
cross-mechanism transfer fails: prompt-induced$\to$fine-tuned recovers $\approx\!0\%$ (a clean failure) and
fine-tuned$\to$prompt only partially (low and noisy), against $+96$--$104\%$ within each mechanism. The two
locks gate capability through \emph{different} directions. The RL-induced axis is also
near-orthogonal to the fine-tuned axis ($|\cos|\!=\!0.17$--$0.26$ across three families, versus $0.06$--$0.13$
for prompt-induced and a $\approx\!0.03$ random floor), so each induction mechanism carries its \emph{own} axis
($\S$\ref{sec:rl}). This refines our ``one operation, two regimes'' claim:
the \emph{operation} (set the coordinate to the honest reference) is shared, but the \emph{direction} is
mechanism-specific---in the wild the contrast must come from the same mechanism that does the sandbagging,
not an arbitrary sandbagging probe. It is an honest boundary on generality, parallel to the
circuit-breaking boundary below.

\smallskip\noindent\textbf{And cheap to fit: a few captures suffice.} Far fewer than our $\sim\!40$-capture
default are needed. Re-fitting the direction from only $m$ examples per class
(we write $m$ for the data budget---captures per class---reserving $k$ for the circuit budget)
(diff-in-means of $m$ honest vs.\ $m$ sandbag captures, AL-selected layers fixed), $m\!=\!2$ already
recovers $+97\%$ of the gap and $m\!=\!5$ recovers $+100\%$, on both Qwen2.5-1.5B and Llama-3-8B, against
$\approx\!0\%$ for a random direction (Table~\ref{tab:fewshot}, appendix). This holds \emph{end-to-end}: running the whole pipeline---AL
circuit \emph{selection} as well as direction and reference---from $m$ paired captures saturates by
$m\!=\!5$ on both models. Strikingly, few-shot AL lands on a
\emph{different but equally effective} circuit set ($\approx\!40\%$ overlap at $m\!=\!5$), reflecting
redundancy across the superposed circuits rather than unstable selection---given the full data, AL is
stable across seeds ($\S$\ref{sec:revert}). A handful of paired captures---never the $160$--$240$
lock-training questions---thus suffices to fit \emph{and} place the withholding axis.

\smallskip\noindent\textbf{The same axis also detects.}\label{sec:detect} The coordinate $s\!=\!h^\top v$
flags, from one forward pass, whether the model is sandbagging. Within-dataset AUROC $\approx\!1$ is near-tautological
($v$ \emph{is} the contrast); the informative test is transfer, where a contrast fit on ARC scores WMDP at AUROC
$\approx\!1.0$ (TPR $1.0$ at $5\%$ FPR; ARC$\leftrightarrow$WMDP, Qwen2.5-1.5B/Llama-3-8B), beating output-confidence
($0.25$--$0.82$)---but fails across mechanisms ($0.42$, seeds $0.0$--$1.0$). One direction is thus both monitor and
actuator within a mechanism.

\subsection{Circuit-Breaking Defeats Grafting}
\label{sec:cb}
\citet{hofstatter2025elicitation} introduce a \emph{harder} organism: circuit-breaking
\citep{zou2024circuit} applied \emph{on top of} the password-lock, designed to be robust to
elicitation. We build it (representation rerouting, RR) and re-test (Figure~\ref{fig:cb}, appendix):
\textbf{circuit-breaking defeats grafting.} On the circuit-broken organism ($n{=}4$ seeds) grafting recovers
only $-11\%$, versus $+99\%$ for the identical graft on the plain lock; deactivation, random patching,
and even fine-tuning all fail ($\le\!19\%$). Rerouting the no-password representations breaks the linear
contrast, so a fixed coordinate no longer reaches the unlocked state.

\paragraph{Gated, not destroyed---and no fixed map undoes it.} RR does not \emph{erase} the capability:
with-password accuracy is within $0.04$ of the base ceiling, and ablating the RR adapter restores the plain
lock---it is held behind \emph{active} rerouting. Nor does a richer fixed map help: a higher-rank affine
transport (orthogonal Procrustes, per-layer ridge) and a trained universal soft-prompt all fail ($\le\!2\%$;
Figure~\ref{fig:transport}, appendix). Yet the ridge map drives the reconstruction residual to $0.14$---and
to $0.01$ \emph{along the gating direction the lock reads}, matching even the gate coordinate while
recovering nothing. \textbf{Reconstructing the geometry is not functional elicitation:} the reroute
re-applies every forward pass, so any fixed upstream edit is re-broken downstream. We claim only that the
\emph{fixed} interventions we test are re-broken; stronger per-input embedding attacks do break circuit
breakers for jailbreak \citep{schwinn2024revisiting,bailey2024obfuscated}.

\paragraph{The released organism confirms it---and a forced-choice metric alone would not.} We close the
released-vs-reconstructed gap by running the \emph{same} pipeline on Hofst\"atter et al.'s \emph{own} released
circuit-broken organism (as in $\S$\ref{sec:revert} for their password locks). Here the forced-choice metric is
\emph{misleading}: grafting appears to recover ($+89\%$ at the answer logit, and $+94$--$99\%$ even when the answer
is \emph{generated}), yet under free decoding the generated text is incoherent (coherence $0.45$ versus
$\approx\!1.0$ with the password), so genuine coherent recovery is only $+35\%$ ($n{=}3$). Representation
rerouting corrupts \emph{generation}, exactly its design target, which a single-token readout never engages. A
\emph{static} lock shows no such gap on the same harness: on the released plain password locks, grafted generation
recovers \emph{and} stays fully coherent (coherence $\approx\!1.0$, equal to the with-password topline, versus
$0.45$ for the circuit-broken organism at the same $+97\%$ raw recovery); and free-decoding the MCQA answer recovers
as well as the forced-choice readout ($+93$--$101\%$ across four models, $n{=}3$). Coherence is
uninformative on the MCQA organisms themselves---they emit a bare letter then loop, a format artifact
rather than rerouting---so the axis is exercised here. \textbf{Coherent generative
evaluation is what separates a static lock from an active defense}---and it is why a forced-choice number on the
released CB organism overstates elicitation.

\section{Discussion}
\paragraph{Refining ``steering fails.''} The negative verdict of \citet{hofstatter2025elicitation} is
sensitive to the steering \emph{operation} (set-to-reference vs.\ additive nudging) and \emph{selection}
(AL-minimal superposed circuits vs.\ hand-chosen): with both right, steering matches fine-tuning. The
interesting question is then not \emph{whether} steering works but \emph{when}.

\paragraph{The reference is the crux---and the key to the wild.} The technique is invariant across cases;
only the reference $t^\star$ changes (an honest prompt; the password-on captures; in the wild, a few
elicited examples or a reference model). The genuinely hard, case-dependent part is \emph{obtaining} a
trustworthy honest reference when a model never behaves honestly unprompted---the same obstacle every
elicitation method faces. Our method does not solve reference-acquisition, but it needs only a sliver of
honest signal, which it \emph{amplifies} into full, cheap, inference-time elicitation while
\emph{localising} where the suppression is gated. What content-generality buys is the \emph{task} axis: the
reference need not come from the target task ($\S$\ref{sec:transfer}), so it can be built on a safe one and
applied to a dangerous one---a handful of unlock captures taken on ARC and grafted on WMDP recover
$\sim\!90\%$ of the dangerous-capability gap, where a \emph{fully password-free} reference (appendix)
recovers only $\sim\!{+}28\%$. What it does not buy is the \emph{mechanism} axis: cross-mechanism transfer
fails ($\approx\!0\%$), so an auditor cannot borrow an easier reference from a different mechanism. Two
conditions must therefore hold for sandbagging in the wild: a cheap
\emph{same-mechanism} honest reference, and a low-rank, linearly contrastable suppression.

\paragraph{Limitations.}\label{sec:lim}
Our claims assume a \emph{static} lock---one a fixed edit inverts. Circuit-breaking ($\S$\ref{sec:cb}) is
the boundary: it reroutes every forward pass, defeating grafting and every fixed map (the capability
\emph{gated, not destroyed}). \emph{Task scope:} our multi-model results are multiple-choice; the code
results are scoped in $\S$\ref{sec:revert}, and open-ended generation remains
untested (a harder-task sensitivity \citet{roger2025fuzzing} also reports). \emph{Induction scope:} RL-induced sandbagging is
graft-invertible at the decision token ($\S$\ref{sec:rl}), but a free-generation variant distributes the
signal beyond it, and whether \emph{naturally-occurring} sandbagging is static is untested.
Finally, we \emph{match}, not beat,
fine-tuning---our value is forward-pass cost and gate localisation---and AL is greedy single-step
(batched/multi-direction acquisition is future work).

\section{Conclusion}
In the organisms we study, sandbagging lives in a few AL-selected superposed circuits, and one
operation---grafting their coordinate to the honest reference---reverts it, matching fine-tuning
at forward-pass cost with the gate localised. The open problems are an honest
reference in the wild and a defence: defeating circuit-breaking.

\section*{Ethical Statement}
This work strengthens safety evaluations by exposing hidden capability
(sandbagging), a defensive goal. WMDP-bio, our one hazardous-knowledge set, is
a standard \emph{audit} benchmark used only as a choice-likelihood probe.
Capability-elicitation is dual-use: it could be turned on a deployed model to
surface restricted capability. But grafting needs white-box access to internal
activations, confining misuse to open-weight models, where fine-tuning already
elicits comparably (Table~\ref{tab:baselines}). We release code and model
organisms: weakened models holding no capability their open-weight bases lack.

\bibliography{references}

@misc{vanderweij2024sandbagging,
  title={AI Sandbagging: Language Models can Strategically Underperform on Evaluations},
  author={van der Weij, Teun and Hofst{\"a}tter, Felix and Jaffe, Ollie and Brown, Samuel F. and Ward, Francis Rhys},
  year={2024}, eprint={2406.07358}, archivePrefix={arXiv}}

@inproceedings{greenblatt2024password,
  title={Stress-Testing Capability Elicitation With Password-Locked Models},
  author={Greenblatt, Ryan and Roger, Fabien and Krasheninnikov, Dmitrii and Krueger, David},
  booktitle={Advances in Neural Information Processing Systems (NeurIPS)},
  year={2024}}

@misc{tice2024noise,
  title={Noise Injection Reveals Hidden Capabilities of Sandbagging Language Models},
  author={Tice, Cameron and Kreer, Philipp Alexander and Helm-Burger, Nathan and Shahani, Prithviraj Singh and Ryzhenkov, Fedor and Haimes, Jacob and Hofst{\"a}tter, Felix and van der Weij, Teun},
  year={2024}, eprint={2412.01784}, archivePrefix={arXiv}}

@misc{hofstatter2025elicitation,
  title={The Elicitation Game: Evaluating Capability Elicitation Techniques},
  author={Hofst{\"a}tter, Felix and van der Weij, Teun and Teoh, Jayden and Bartsch, Henning and Ward, Francis Rhys},
  year={2025}, eprint={2502.02180}, archivePrefix={arXiv}}

@article{che2025tampering,
  title={Model Tampering Attacks Enable More Rigorous Evaluations of LLM Capabilities},
  author={Che, Zora and Casper, Stephen and Kirk, Robert and Satheesh, Anirudh and Slocum, Stewart and McKinney, Lev E. and Gandikota, Rohit and Ewart, Aidan and Rosati, Domenic and Wu, Zichu and Cai, Zikui and Chughtai, Bilal and Gal, Yarin and Huang, Furong and Hadfield-Menell, Dylan},
  journal={Transactions on Machine Learning Research (TMLR)},
  year={2025}}

@misc{zou2024circuit,
  title={Improving Alignment and Robustness with Circuit Breakers},
  author={Zou, Andy and Phan, Long and Wang, Justin and Duenas, Derek and Lin, Maxwell and Andriushchenko, Maksym and Wang, Rowan and Kolter, Zico and Fredrikson, Matt and Hendrycks, Dan},
  year={2024}, eprint={2406.04313}, archivePrefix={arXiv}}

@inproceedings{wu2025axbench,
  title={{AxBench}: Steering {LLM}s? Even Simple Baselines Outperform Sparse Autoencoders},
  author={Wu, Zhengxuan and Arora, Aryaman and Geiger, Atticus and Wang, Zheng and Huang, Jing and Jurafsky, Dan and Manning, Christopher D. and Potts, Christopher},
  booktitle={International Conference on Machine Learning (ICML)},
  year={2025}}

@misc{hua2025steering,
  title={Steering Evaluation-Aware Language Models to Act Like They Are Deployed},
  author={Hua, Tim Tian and Qin, Andrew and Marks, Samuel and Nanda, Neel},
  year={2025}, eprint={2510.20487}, archivePrefix={arXiv}}

@misc{schwinn2024revisiting,
  title={Revisiting the Robust Alignment of Circuit Breakers},
  author={Schwinn, Leo and Geisler, Simon},
  year={2024}, eprint={2407.15902}, archivePrefix={arXiv}}

@misc{bailey2024obfuscated,
  title={Obfuscated Activations Bypass {LLM} Latent-Space Defenses},
  author={Bailey, Luke and Serrano, Alex and Sheshadri, Abhay and Seleznyov, Mikhail and Taylor, Jordan and Jenner, Erik and Hilton, Jacob and Casper, Stephen and Guestrin, Carlos and Emmons, Scott},
  year={2024}, eprint={2412.09565}, archivePrefix={arXiv}}

@misc{he2025locket,
  title={Locket: Robust Feature-Locking Technique for Language Models},
  author={He, Lipeng and Duddu, Vasisht and Asokan, N.},
  year={2025}, eprint={2510.12117}, archivePrefix={arXiv}}

@misc{roger2025fuzzing,
  title={Fuzzing {LLM}s Sometimes Makes Them Reveal Their Secrets},
  author={Roger, Fabien},
  year={2025},
  note={AI Alignment Forum. \url{https://www.alignmentforum.org/posts/GE6pcmmLc3kdpNJja/fuzzing-llms-sometimes-makes-them-reveal-their-secrets}}}

@inproceedings{vig2020causal,
  title={Investigating Gender Bias in Language Models Using Causal Mediation Analysis},
  author={Vig, Jesse and Gehrmann, Sebastian and Belinkov, Yonatan and Qian, Sharon and Nevo, Daniel and Singer, Yaron and Shieber, Stuart},
  booktitle={Advances in Neural Information Processing Systems (NeurIPS)},
  year={2020}}

@inproceedings{meng2022rome,
  title={Locating and Editing Factual Associations in {GPT}},
  author={Meng, Kevin and Bau, David and Andonian, Alex and Belinkov, Yonatan},
  booktitle={Advances in Neural Information Processing Systems (NeurIPS)},
  year={2022}}

@inproceedings{geiger2021causal,
  title={Causal Abstractions of Neural Networks},
  author={Geiger, Atticus and Lu, Hanson and Icard, Thomas and Potts, Christopher},
  booktitle={Advances in Neural Information Processing Systems (NeurIPS)},
  year={2021}}

@inproceedings{zhang2024patching,
  title={Towards Best Practices of Activation Patching in Language Models: Metrics and Methods},
  author={Zhang, Fred and Nanda, Neel},
  booktitle={International Conference on Learning Representations (ICLR)},
  year={2024}}

@inproceedings{marks2024geometry,
  title={The Geometry of Truth: Emergent Linear Structure in Large Language Model Representations of True/False Datasets},
  author={Marks, Samuel and Tegmark, Max},
  booktitle={Conference on Language Modeling (COLM)},
  year={2024}}

@inproceedings{parvaneh2022alphamix,
  title={Active Learning by Feature Mixing},
  author={Parvaneh, Amin and Abbasnejad, Ehsan and Teney, Damien and Haffari, Reza and van den Hengel, Anton and Shi, Javen Qinfeng},
  booktitle={IEEE/CVF Conference on Computer Vision and Pattern Recognition (CVPR)},
  year={2022}
}

@inproceedings{wang2022ioi,
  title={Interpretability in the Wild: a Circuit for Indirect Object Identification in GPT-2 small},
  author={Wang, Kevin and Variengien, Alexandre and Conmy, Arthur and Shlegeris, Buck and Steinhardt, Jacob},
  booktitle={International Conference on Learning Representations (ICLR)},
  year={2023}
}

@inproceedings{arditi2024refusal,
  title={Refusal in Language Models Is Mediated by a Single Direction},
  author={Arditi, Andy and Obeso, Oscar and Syed, Aaquib and Paleka, Daniel and Panickssery, Nina and Gurnee, Wes and Nanda, Neel},
  booktitle={Advances in Neural Information Processing Systems (NeurIPS)},
  year={2024}
}

@inproceedings{conmy2023acdc,
  title={Towards Automated Circuit Discovery for Mechanistic Interpretability},
  author={Conmy, Arthur and Mavor-Parker, Augustine N. and Lynch, Aengus and Heimersheim, Stefan and Garriga-Alonso, Adri{\`a}},
  booktitle={Advances in Neural Information Processing Systems (NeurIPS)},
  year={2023}
}

@inproceedings{marks2024sparse,
  title={Sparse Feature Circuits: Discovering and Editing Interpretable Causal Graphs in Language Models},
  author={Marks, Samuel and Rager, Can and Michaud, Eric J. and Belinkov, Yonatan and Bau, David and Mueller, Aaron},
  booktitle={International Conference on Learning Representations (ICLR)},
  year={2025}
}

@inproceedings{kirsch2019batchbald,
  title={BatchBALD: Efficient and Diverse Batch Acquisition for Deep Bayesian Active Learning},
  author={Kirsch, Andreas and van Amersfoort, Joost and Gal, Yarin},
  booktitle={Advances in Neural Information Processing Systems (NeurIPS)},
  year={2019}
}

@inproceedings{ash2020badge,
  title={Deep Batch Active Learning by Diverse, Uncertain Gradient Lower Bounds},
  author={Ash, Jordan T. and Zhang, Chicheng and Krishnamurthy, Akshay and Langford, John and Agarwal, Alekh},
  booktitle={International Conference on Learning Representations (ICLR)},
  year={2020}
}

@inproceedings{bickfordsmith2023epig,
  title={Prediction-Oriented Bayesian Active Learning},
  author={Bickford Smith, Freddie and Kirsch, Andreas and Farquhar, Sebastian and Gal, Yarin and Foster, Adam and Rainforth, Tom},
  booktitle={International Conference on Artificial Intelligence and Statistics (AISTATS)},
  year={2023}
}

@inproceedings{sener2018coreset,
  title={Active Learning for Convolutional Neural Networks: A Core-Set Approach},
  author={Sener, Ozan and Savarese, Silvio},
  booktitle={International Conference on Learning Representations (ICLR)},
  year={2018}
}

@misc{elhage2022superposition,
  title={Toy Models of Superposition},
  author={Elhage, Nelson and Hume, Tristan and Olsson, Catherine and Schiefer, Nicholas and Henighan, Tom and Kravec, Shauna and Hatfield-Dodds, Zac and Lasenby, Robert and Drain, Dawn and Chen, Carol and Grosse, Roger and McCandlish, Sam and Kaplan, Jared and Amodei, Dario and Wattenberg, Martin and Olah, Christopher},
  year={2022}, eprint={2209.10652}, archivePrefix={arXiv}}

@inproceedings{cunningham2023sae,
  title={Sparse Autoencoders Find Highly Interpretable Features in Language Models},
  author={Cunningham, Hoagy and Ewart, Aidan and Riggs, Logan and Huben, Robert and Sharkey, Lee},
  booktitle={International Conference on Learning Representations (ICLR)}, year={2024}}

@misc{korznikov2025rogue,
  title={The Rogue Scalpel: Activation Steering Compromises LLM Safety},
  author={Korznikov and Galichin and Dontsov and Rogov and Oseledets, Ivan and Tutubalina, Elena},
  year={2025}, eprint={2509.22067}, archivePrefix={arXiv}}

@misc{zou2023repe,
  title={Representation Engineering: A Top-Down Approach to AI Transparency},
  author={Zou, Andy and Phan, Long and Chen, Sarah and Campbell, James and Guo, Phillip and Ren, Richard and Pan, Alexander and Yin, Xuwang and Mazeika, Mantas and Dombrowski, Ann-Kathrin and Goel, Shashwat and Li, Nathaniel and Byun, Michael J. and Wang, Zifan and Mallen, Alex and Basart, Steven and Koyejo, Sanmi and Song, Dawn and Fredrikson, Matt and Kolter, J. Zico and Hendrycks, Dan},
  year={2023}, eprint={2310.01405}, archivePrefix={arXiv}}

@inproceedings{rimsky2023caa,
  title={Steering Llama 2 via Contrastive Activation Addition},
  author={Rimsky, Nina and Gabrieli, Nick and Schulz, Julian and Tong, Meg and Hubinger, Evan and Turner, Alexander Matt},
  booktitle={Annual Meeting of the Association for Computational Linguistics (ACL)},
  year={2024}}

@misc{marshall2024refusal,
  title={Refusal in {LLM}s is an Affine Function},
  author={Marshall, Thomas and Scherlis, Adam and Belrose, Nora},
  year={2024}, eprint={2411.09003}, archivePrefix={arXiv}}

@inproceedings{makelov2024subspace,
  title={Is This the Subspace You Are Looking For? An Interpretability Illusion for Subspace Activation Patching},
  author={Makelov, Aleksandar and Lange, Georg and Nanda, Neel},
  booktitle={International Conference on Learning Representations (ICLR)},
  year={2024}}

@inproceedings{geiger2024finding,
  title={Finding Alignments Between Interpretable Causal Variables and Distributed Neural Representations},
  author={Geiger, Atticus and Wu, Zhengxuan and Potts, Christopher and Icard, Thomas and Goodman, Noah},
  booktitle={Causal Learning and Reasoning (CLeaR)},
  year={2024}}

@inproceedings{rodriguez2024controlling,
  title={Controlling Language and Diffusion Models by Transporting Activations},
  author={Rodriguez, Pau and Blaas, Arno and Klein, Michal and Zappella, Luca and Apostoloff, Nicholas and Cuturi, Marco and Suau, Xavier},
  booktitle={International Conference on Learning Representations (ICLR)},
  year={2025}}

@misc{clark2018arc,
  title={Think you have Solved Question Answering? Try {ARC}, the {AI2} Reasoning Challenge},
  author={Clark, Peter and Cowhey, Isaac and Etzioni, Oren and Khot, Tushar and Sabharwal, Ashish and Schoenick, Carissa and Tafjord, Oyvind},
  year={2018}, eprint={1803.05457}, archivePrefix={arXiv}}

@inproceedings{li2024wmdp,
  title={The {WMDP} Benchmark: Measuring and Reducing Malicious Use with Unlearning},
  author={Li, Nathaniel and Pan, Alexander and Gopal, Anjali and Yue, Summer and Berrios, Daniel and Gatti, Alice and Li, Justin D. and Dombrowski, Ann-Kathrin and Goel, Shashwat and Mukobi, Gabriel and others},
  booktitle={International Conference on Machine Learning (ICML)},
  year={2024}}

@inproceedings{mihaylov2018obqa,
  title={Can a Suit of Armor Conduct Electricity? A New Dataset for Open Book Question Answering},
  author={Mihaylov, Todor and Clark, Peter and Khot, Tushar and Sabharwal, Ashish},
  booktitle={Empirical Methods in Natural Language Processing (EMNLP)},
  year={2018}}

@misc{austin2021mbpp,
  title={Program Synthesis with Large Language Models},
  author={Austin, Jacob and Odena, Augustus and Nye, Maxwell and Bosma, Maarten and Michalewski, Henryk and Dohan, David and Jiang, Ellen and Cai, Carrie and Terry, Michael and Le, Quoc and Sutton, Charles},
  year={2021}, eprint={2108.07732}, archivePrefix={arXiv}}

\appendix
\setcounter{figure}{0}
\setcounter{table}{0}
\setcounter{algorithm}{0}
\renewcommand{\thefigure}{S\arabic{figure}}
\renewcommand{\thetable}{S\arabic{table}}
\renewcommand{\thealgorithm}{S\arabic{algorithm}}

\section{Datasets and Hyperparameters}
MCQA tasks: ARC-Easy (primary), with WMDP-bio/cyber and OpenBookQA for cross-dataset transfer; accuracy by
choice-letter log-likelihood, across eleven models ($1.5$--$32$B, three architecture lineages). Prompt-induced sandbagging via
a forceful system instruction (an ``opposite-game'' variant for models that resist a plain instruction);
password-lock by LoRA ($r{=}16$) to answer correctly only with a password token. Contrast directions:
diff-in-means of honest/with-password vs.\ sandbag/no-password last-token residuals over the mid-to-late
band. AL budget $\le5$ circuits; greedy-KL uses $\mathrm{KL}(\text{honest}\,\|\,\text{grafted})$ on TRAIN.
Grafting reference $t^\star=\mu^{\mathrm{ref}}{}^{\!\top}v$. SAE: $8\times$ overcomplete, $L_1$ penalty,
decoder unit-normalised. Recovery on held-out TEST ($n{=}100$), means over $3$--$6$ seeds, with
random-direction and honest-model controls.
\emph{Code generation} ($\S$\ref{sec:revert}): Qwen2.5-Coder-1.5B-Instruct on MBPP-sanitized
(train$\to$lock, validation$\to$AL, test$\to$eval); a Greenblatt-style password lock with the reference
solution as the with-password target and a weak model's (TinyLlama-1.1B) completion as the no-password
target; contrast and reference taken over teacher-forced \emph{completion} tokens (per-layer scalar
$t^\star$); graft applied at all positions during decoding; accuracy is execution-based pass@1 from MBPP
\texttt{test\_list} asserts; AL uses label-free greedy-KL over completion positions (no execution in the
loop). \emph{Third-family stress test:} StarCoder2-7B ($n{=}1$, gap $0.38$) recovers $+82\%$ under
grafting and $+79\%$ under deactivation, but with a with-password fluency drop of $0.12$, well above the
$\le0.04$ of the three code-specialized organisms in Table~\ref{tab:code}; we report it as an
indicative single-seed check, not as evidence.
\emph{Password-free transfer} (Discussion sanity check; Qwen2.5-1.5B fine-tuned two-way \{ARC,WMDP\} lock,
$n{=}3$): the contrast direction is estimated on the \emph{safe} task (ARC) \emph{without} the password or
any labels---the gradient of $\log p(\text{correct})$ w.r.t.\ the last-token residual (the few-shot
estimator of $\S$\ref{sec:transfer}, $\cos$ to the privileged axis $\approx\!0.05$)---then grafted on the
\emph{danger} task (WMDP) held-out test. We report it against the privileged-ARC ceiling ($\sim\!100\%$), a
random direction, and a calibrated-random control (direction matters iff password-free $\gg$ calibrated
random); recovery is $\sim\!{+}28\%$, above the controls but well below the ceiling.
\emph{Organism training.} Password locks are LoRA adapters ($r{=}16$, $\alpha{=}32$, dropout $0.05$,
applied to the attention projections $q,k,v,o$) trained for $3$ epochs with AdamW at learning rate
$2\times10^{-4}$. The RL organism is GRPO for $200$ steps at learning rate $10^{-4}$, batch $16$, group
size $6$, sampling temperature $1.8$ and entropy bonus $0.03$, with no KL anchor to the reference policy
($\beta{=}0$, which we found fights the lock). The contrast band spans layers $\lceil0.34L\rceil$ through
$L$ of an $L$-layer model.
\emph{Computing infrastructure.} Most experiments run on a local workstation with two NVIDIA GeForce RTX
3090 GPUs ($24$\,GB each), an Intel Core i7-12700K, and $62$\,GB of RAM, under Debian 13 (Linux 6.12),
running Python~$3.12$ with PyTorch~$2.12$ (CUDA~$13$), HuggingFace \texttt{transformers}~$5.9.0$,
\texttt{datasets}~$5.0.0$ and \texttt{peft}~$0.19.1$, \texttt{scikit-learn}~$1.8.0$,
\texttt{numpy}~$2.3.3$ and \texttt{matplotlib}~$3.11.0$. The organisms that do not fit in $24$\,GB
(Qwen2.5-32B and Gemma-2-27B) run instead on rented NVIDIA L40S, A100 and H100 instances on Lambda, under
Python~$3.10$ with the CUDA image's own PyTorch build. Both stacks are pinned in the released
\texttt{requirements-local.txt} and \texttt{requirements-lambda.txt}. For the released-organism
replication ($\S$\ref{sec:revert}) we instead matched the pinned stack of \citet{hofstatter2025elicitation}
(\texttt{torch}~$2.3.1$, \texttt{transformers}~$4.44.2$, \texttt{peft}~$0.11.1$).
Seeds are set through \texttt{torch.manual\_seed} and
\texttt{numpy.random.seed} from a \texttt{-\/-seed} flag, which fixes the data split; GPU training is not
bitwise deterministic, so per-seed draws vary run to run and all reported figures are means over the
stated number of seeds.
Per-model JSON outputs and code accompany the paper under a BSD 3-clause license.

\section{Additional Figures and Tables}

\begin{table}[h]
\centering\small\setlength{\tabcolsep}{8pt}
\begin{tabular}{@{}ccc@{}}
\toprule
$m$ / class & $|\cos|$ to full axis & recovery (\% gap) \\
\midrule
$1$        & $0.85$ & $+91\pm19$ \\
$2$        & $0.90$ & $+97\pm6$  \\
$5$        & $0.97$ & $+100\pm6$ \\
$10$       & $0.98$ & $+101\pm6$ \\
$50$       & $1.00$ & $+99\pm3$  \\
\midrule
random dir & ---    & $+0$       \\
\bottomrule
\end{tabular}
\caption{Fitting the withholding axis from $m$ paired honest/sandbag captures (Qwen2.5-1.5B, $4$ seeds; AL
circuits fixed). Just $2$--$5$ captures recover the full $\sim\!40$-capture axis ($|\cos|\,0.90$--$0.97$)
and $+97$--$100\%$ of the gap; a random direction recovers nothing. Llama-3-8B matches ($m{=}5$:
$|\cos|\,0.97$, $+100\%$). Running AL \emph{selection} end-to-end from the same $m$ also saturates by
$m{=}5$ ($+98$--$101\%$, on a different but equally effective circuit set; $\S$\ref{sec:transfer}).}
\label{tab:fewshot}
\end{table}

\begin{figure*}[h]
\centering
\includegraphics[width=\textwidth]{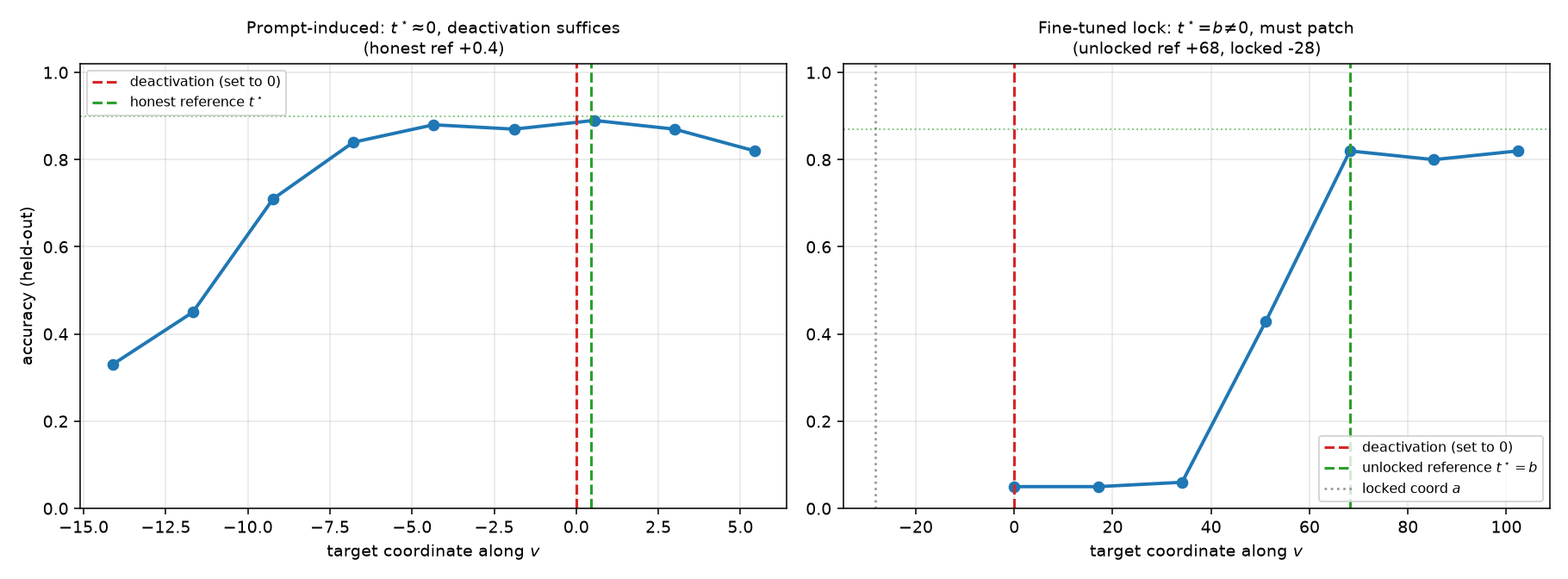}
\caption{One operation, two regimes. Setting the coordinate along $v$ to the honest reference $t^\star$.
\emph{Left, prompt-induced:} the reference is $\approx\!0$, so deactivation lands on the recovered
plateau. \emph{Right, fine-tuned lock:} the reference is $b\!\neq\!0$ and the plateau \emph{excludes}
$0$, so deactivation fails and patching to $b$ is required.}
\label{fig:unification}
\end{figure*}

\begin{figure}[h]
\centering
\includegraphics[width=0.99\columnwidth]{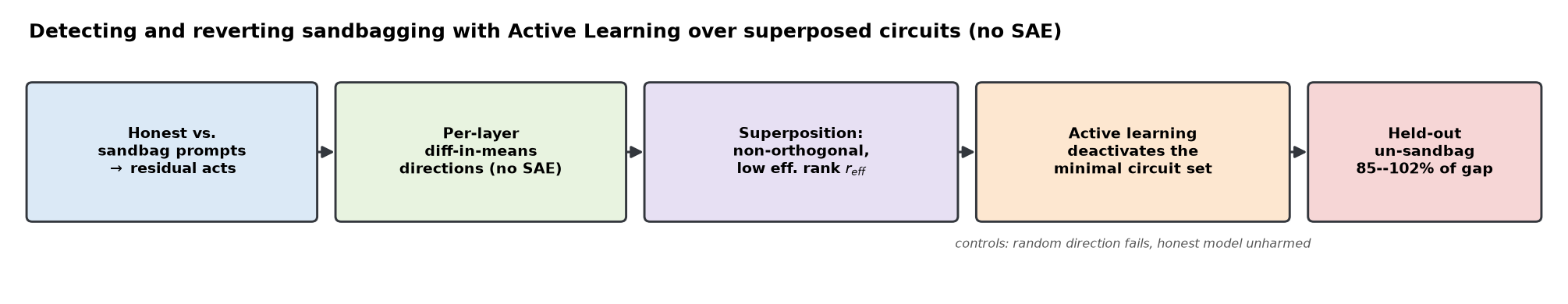}
\caption{Overview. From contrastive (honest vs.\ sandbag) prompts we build per-layer diff-in-means
directions \emph{without an SAE}; these are non-orthogonal and low effective rank---superposed. Active
learning selects the minimal circuits, and we \emph{graft} each to its honest-reference value
(Eq.~\eqref{eq:graft})---deactivation when the reference is $\approx\!0$ (prompt-induced), patching when
it is not (fine-tuned). Recovery is measured on held-out questions with random-direction and
honest-model controls.}
\label{fig:overview}
\end{figure}

\begin{figure}[h]
\centering
\includegraphics[width=0.92\columnwidth]{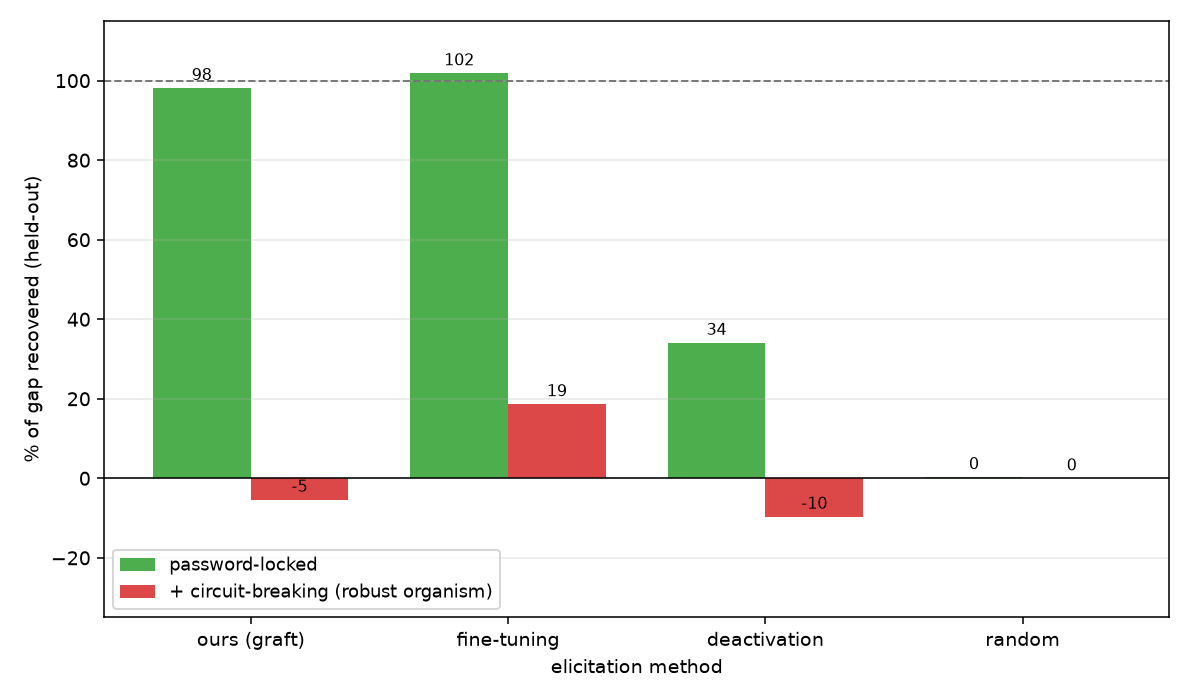}
\caption{Circuit-breaking \citep{zou2024circuit} defeats grafting (Qwen2.5-1.5B, $n{=}3$). Grafting
matches fine-tuning on the password-lock (green) but fails after circuit-breaking (red); even
fine-tuning is substantially weakened. Circuit-breaking reroutes the contrast our grafting relies on.}
\label{fig:cb}
\end{figure}

\begin{figure}[h]
\centering
\includegraphics[width=0.85\columnwidth]{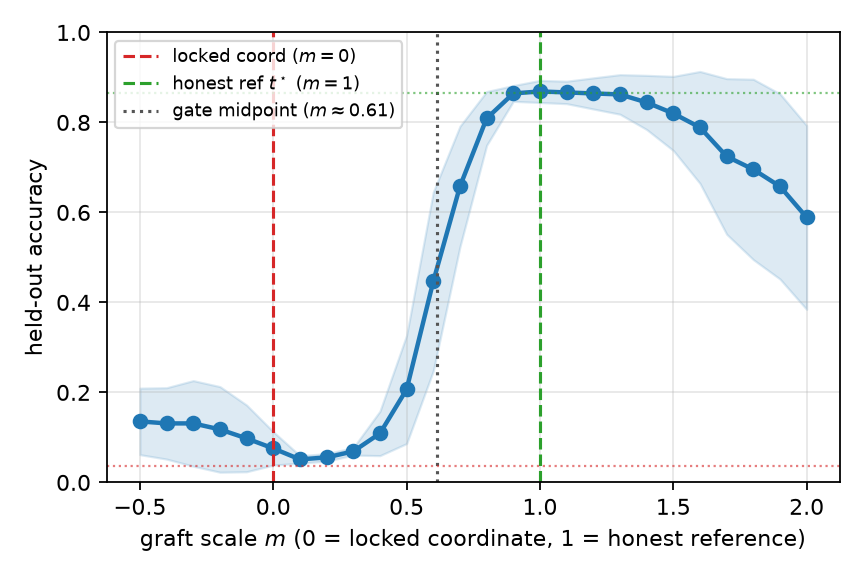}
\caption{The thresholded gate, measured. Sweeping the grafted coordinate from the locked value ($m{=}0$) to
the honest reference ($m{=}1$) along $v$, held-out accuracy is flat-low, steps up at a midpoint
$m\!\approx\!0.61$ ($10$--$90\%$ width $0.32$), and plateaus beyond $t^\star$ (Qwen2.5-1.5B fine-tuned
password-lock, mean$\pm$std over $3$ seeds)---a sharp step, not a linear ramp.}
\label{fig:gatesweep}
\end{figure}

\begin{figure*}[h]
\centering
\includegraphics[width=\textwidth]{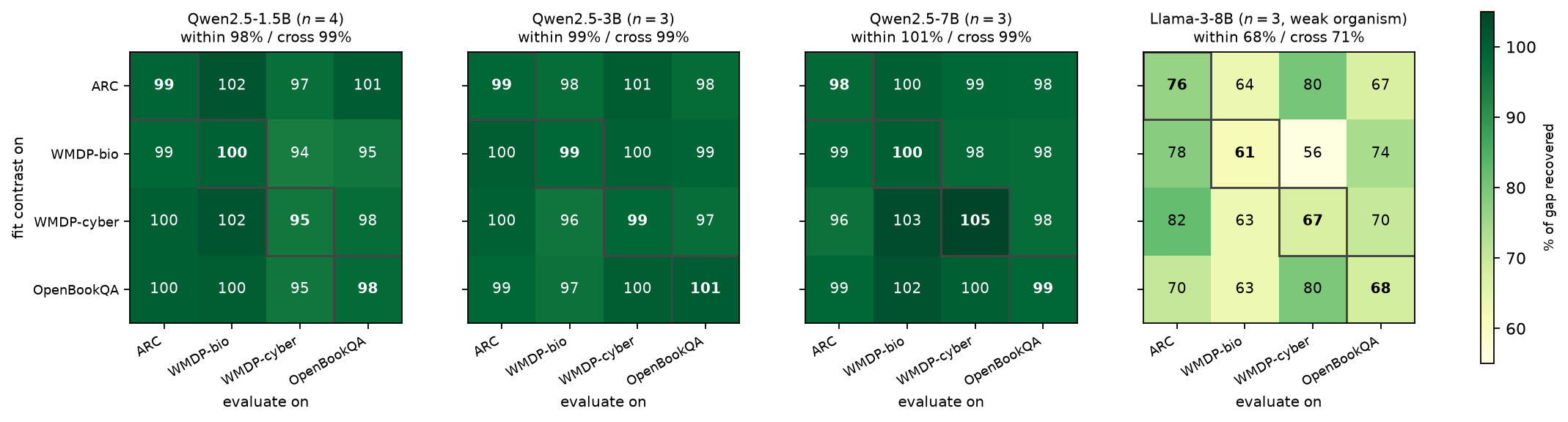}
\caption{Shared four-dataset-lock transfer (the secondary, cross-family check; cf.\ the modular
pairwise matrix, Figure~\ref{fig:pairwise}). One lock mixes all four datasets; contrast fit on the row
dataset, grafted on the column's held-out test. Off-diagonal (cross) tracks the boxed diagonal (within)
throughout: the \emph{Qwen2.5 family is scale-flat} (within/cross $98$/$99$, $99$/$99$, $101$/$99$ at
$1.5$/$3$/$7$B). \emph{Llama-3-8B} is a weaker, high-variance organism with lower absolute recovery, yet its
off-diagonal still tracks its own diagonal ($68$/$71$); Qwen3-4B sits between ($87$/$87$). The lower
$4$B/$8$B levels are thus a family effect, not scale.}
\label{fig:transferheat}
\end{figure*}

\begin{figure*}[h]
\centering
\includegraphics[width=\textwidth]{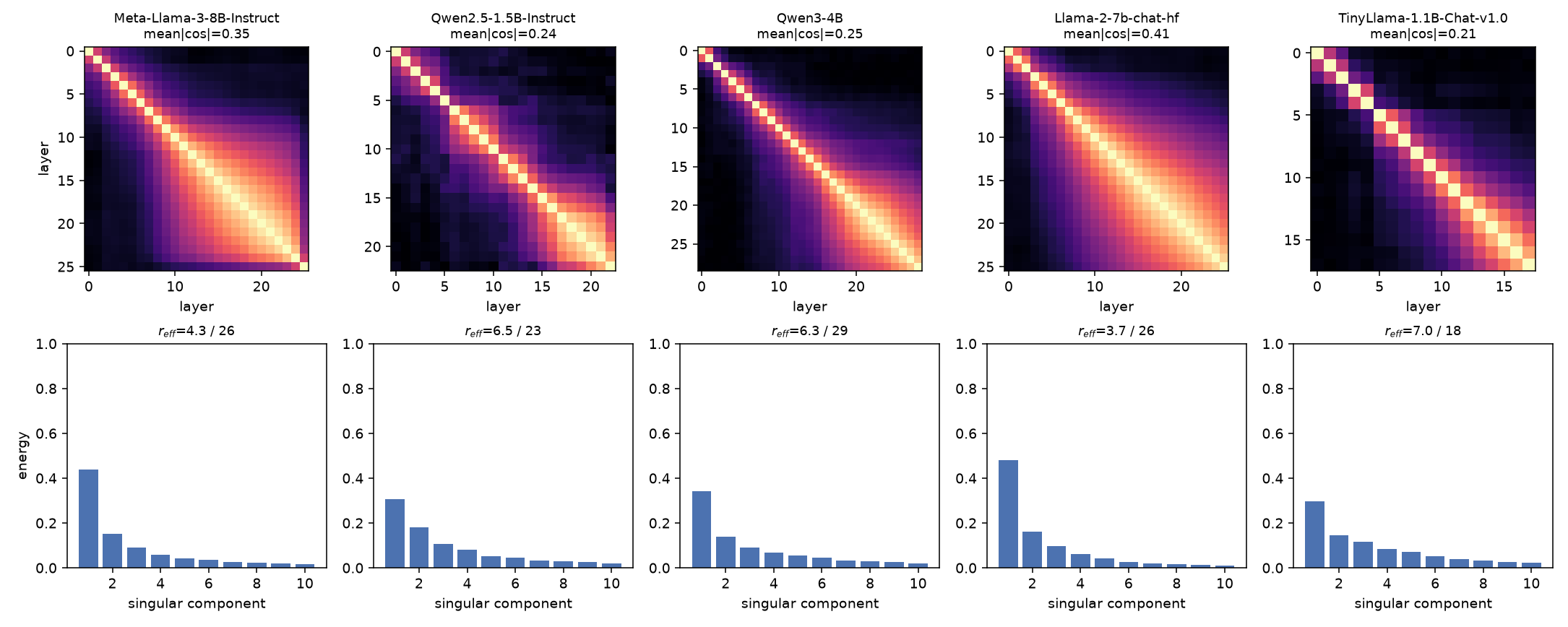}
\caption{Superposition geometry of the sandbagging signal across five models (no SAE). \emph{Top:}
pairwise $|\cos|$ of the per-layer directions---strongly non-orthogonal. \emph{Bottom:} singular-value
energy; $r_{\mathrm{eff}}$ is far below the layer count.}
\label{fig:geom}
\end{figure*}

\begin{algorithm}[h]
\caption{AL over superposed circuits, reference-grafting}
\label{alg:al}
\begin{algorithmic}[1]
\STATE \textbf{Input:} model, TRAIN/TEST, layer band $\mathcal{B}$, budget $K$, acquisition $a(\cdot)$, reference acts
\FOR{$\ell\in\mathcal{B}$}
  \STATE $v_\ell\!\leftarrow\!\mathrm{unit}(\bar h_\ell^{\mathrm{ref}}-\bar h_\ell^{\mathrm{sand}})$;\quad $t^\star_\ell\!\leftarrow\!\mu^{\mathrm{ref}}_\ell{}^{\!\top}v_\ell$
\ENDFOR
\STATE measure superposition: mean $|\cos|$, $r_{\mathrm{eff}}$ \hfill // Eq.~\eqref{eq:pr}
\STATE $S\leftarrow\emptyset$
\WHILE{$|S|<K$}
  \STATE $\ell^\star\!\leftarrow\!\arg\max_{\ell\notin S} a(S\cup\{\ell\})$ on TRAIN under grafting~\eqref{eq:graft}
  \STATE $S\leftarrow S\cup\{\ell^\star\}$
\ENDWHILE
\STATE \textbf{report} TEST accuracy with $S$ grafted; vs.\ random-direction and honest controls
\end{algorithmic}
\end{algorithm}

\begin{table}[h]
\centering\small\setlength{\tabcolsep}{3pt}
\begin{tabular}{@{}lccccc@{}}
\toprule
Model & honest & sandbag & gap & \textbf{ours (graft)} & circuits \\
\midrule
Llama-3-8B                & $0.94$ & $0.04$ & $0.90$ & $\mathbf{+91}$  & $2$ \\
Llama-2-7B                & $0.64$ & $0.22$ & $0.42$ & $\mathbf{+24}$  & $2$ \\
Qwen3-4B$^\ddagger$       & $0.98$ & $0.92$ & $0.06$ & $\mathbf{+67}$  & $3$ \\
Qwen2.5-1.5B$^\ddagger$   & $0.90$ & $0.86$ & $0.04$ & $\mathbf{+100}$ & $1$ \\
TinyLlama-1.1B$^\dagger$  & $0.32$ & $0.18$ & $0.14$ & $\mathbf{+43}$  & $3$ \\
\bottomrule
\end{tabular}
\caption{Prompt-induced sandbagging reverted on five models (\% of gap recovered, held-out). ``gap''$\,{=}\,
\text{honest}{-}\text{sandbag}$; ``ours (graft)''$\,{=}\,(\text{recovered}{-}\text{sandbag})/\text{gap}$
(same metric as Table~\ref{tab:baselines}); ``circuits'' is the number of AL-selected circuits grafted.
Here the unlocked reference is $\approx\!0$, so grafting coincides with deactivation. Recovery is
model-dependent and noisy. $^\ddagger$small gap ($\le\!0.06$); $^\dagger$honest near $4$-way chance
($0.25$)---daggered percentages are indicative only.}
\label{tab:models}
\end{table}

\begin{figure}[h]
\centering
\includegraphics[width=0.99\columnwidth]{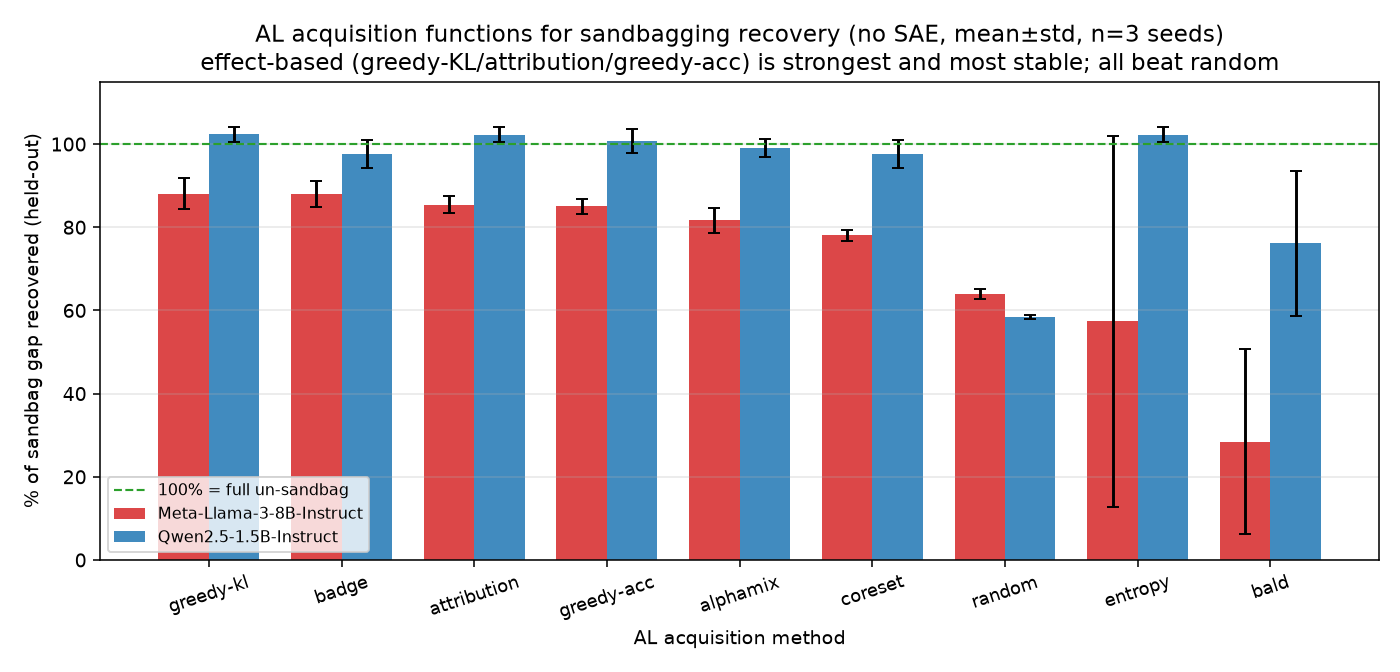}
\caption{Prompt-induced recovery by AL acquisition (mean$\pm$std, $3$ seeds, no SAE). Effect-based AL is
strongest and most stable; all principled methods beat random.}
\label{fig:almethods}
\end{figure}

\begin{table}[h]
\centering\small\setlength{\tabcolsep}{4pt}
\begin{tabular}{@{}lccc@{}}
\toprule
acquisition & Qwen2.5-1.5B & Llama-3-8B & Qwen3-4B \\
\midrule
greedy-KL    & $99$ & $101$ & $100$ \\
attribution  & $98$ & $100$ & $100$ \\
greedy-acc   & $97$ & $100$ & $99$ \\
coreset      & $99$ & $101$ & $99$ \\
badge        & $97$ & $100$ & $99$ \\
alphamix     & $96$ & $100$ & $100$ \\
bald         & $91$ & $97$  & $97$ \\
entropy      & $35$ & $101$ & $68$ \\
\bottomrule
\end{tabular}
\caption{Counterfactual patching of fine-tuned password-locked sandbagging is robust to the AL acquisition
function (\% of gap recovered, held-out; best over $\le5$ circuits per method; Qwen2.5-1.5B/Llama-3-8B
$n{=}3$, Qwen3-4B $n{=}6$). $7/8$ methods recover $90$--$101\%$ (per-method std $1$--$5\%$); entropy is the
unstable exception. With-password accuracy preserved; deactivation and random-patch controls are reported
across all eleven models in Table~\ref{tab:baselines}.}
\label{tab:patch}
\end{table}

\begin{figure}[h]
\centering
\includegraphics[width=0.99\columnwidth]{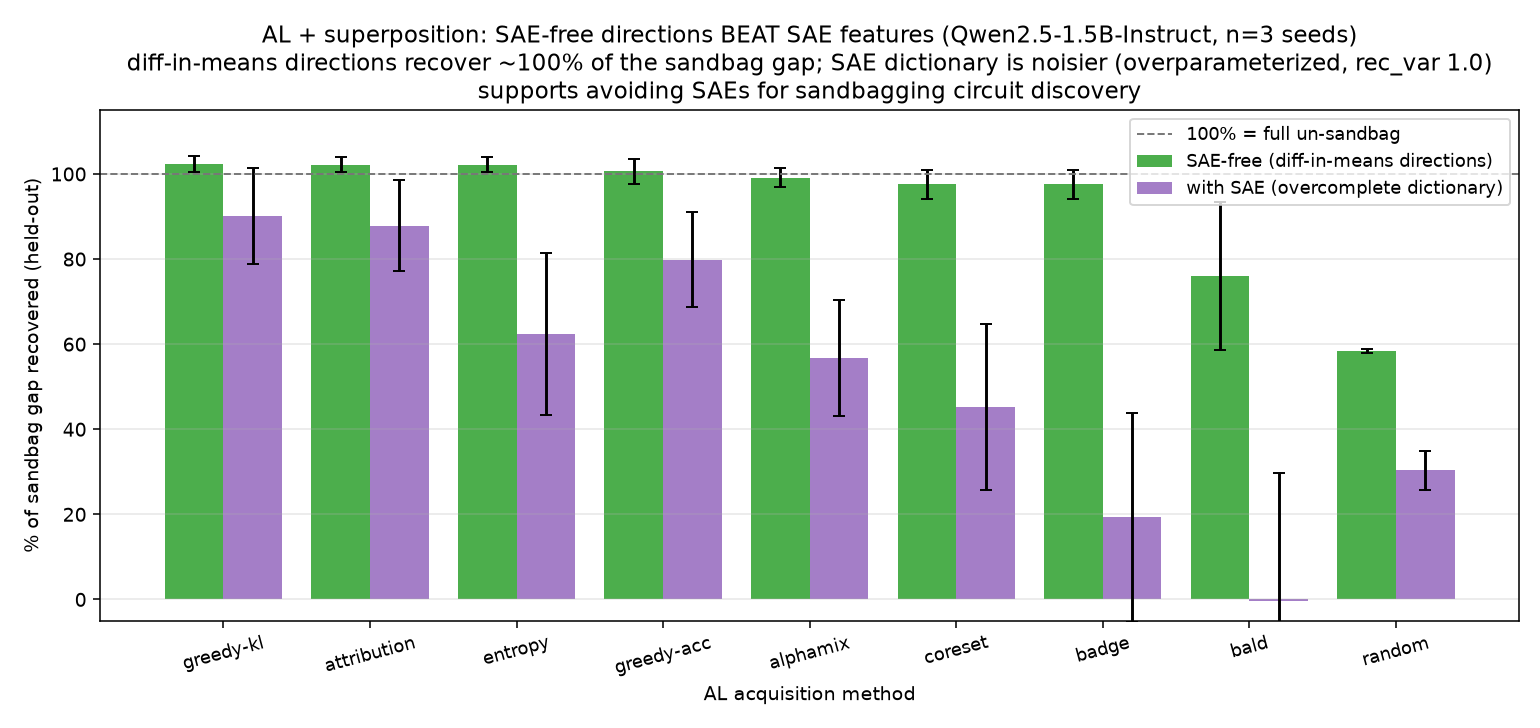}
\caption{SAE-free diff-in-means directions beat an $8\times$ overcomplete SAE dictionary on both models
across acquisition functions ($n{=}3$ seeds).}
\label{fig:sae}
\end{figure}

\begin{figure}[h]
\centering
\includegraphics[width=0.99\columnwidth]{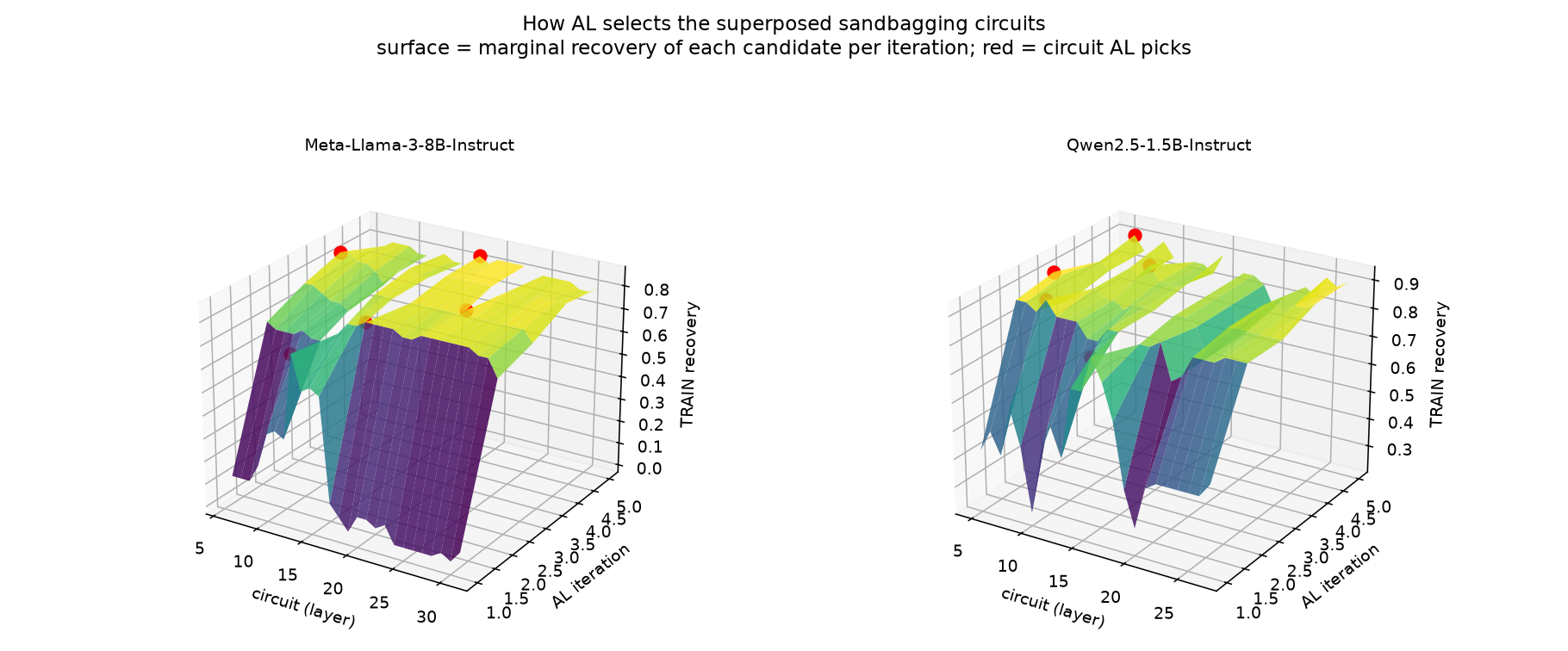}
\caption{How AL selects the superposed circuits. The surface is each candidate circuit's marginal
recovery at each AL iteration; red markers are the circuits AL acquires---a few high-impact directions
within the superposed band.}
\label{fig:traj}
\end{figure}

\begin{figure*}[h]
\centering
\includegraphics[width=\textwidth]{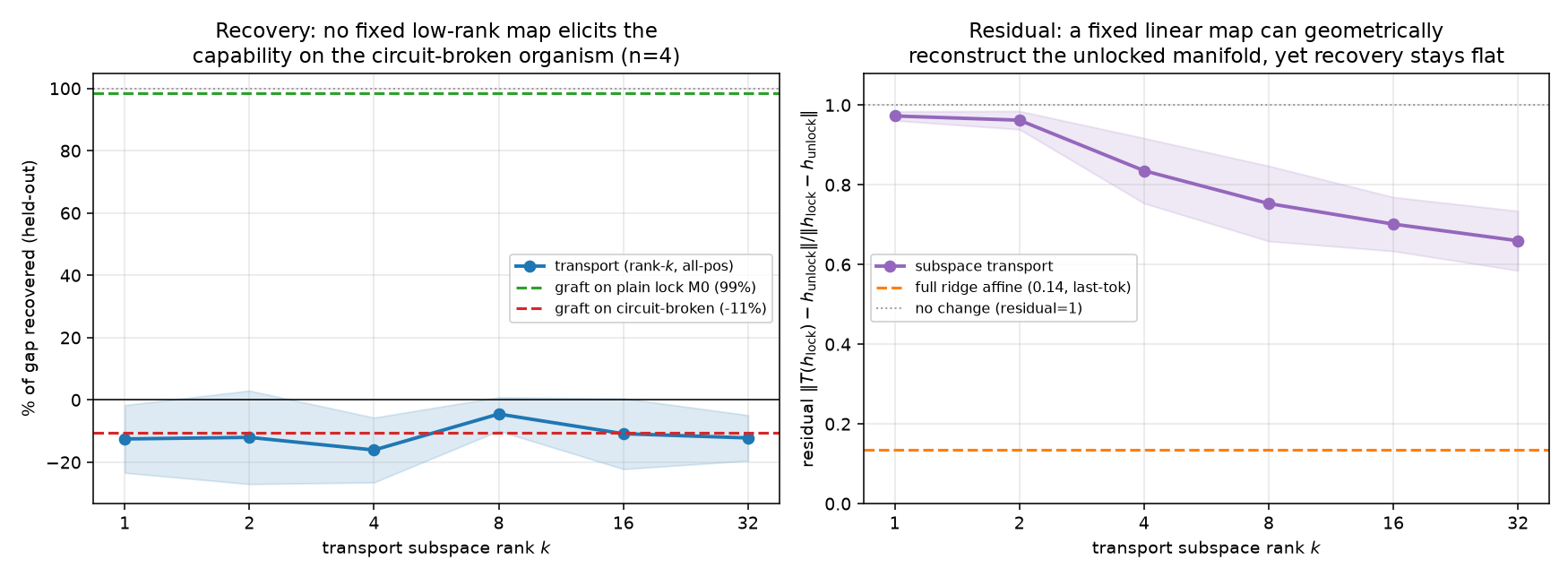}
\caption{Generalising grafting to a fixed transport map does not elicit the circuit-broken organism
($n{=}4$ valid seeds). \emph{Left:} the same graft family recovers $99\%$ on the plain lock (green) but
neither rank-$k$ transport (blue), nor a full ridge affine, nor grafting (red) recovers on the
circuit-broken organism. \emph{Right:} yet the reconstruction residual falls to $0.14$---and to $0.01$
along the gating direction the lock reads---while recovery stays $\approx\!0$: the fixed map matches even
the gate coordinate, so the defence must be an \emph{active} per-forward-pass reroute, not a static
rotation any fixed map could invert.}
\label{fig:transport}
\end{figure*}

\end{document}